\documentclass[11pt]{article}

\usepackage{acl}

\usepackage{arydshln}
\usepackage{latexsym}
\usepackage[T2A,T1]{fontenc}
\usepackage[utf8]{inputenc}
\usepackage{newtxtext,newtxmath}
\usepackage[ukrainian,english]{babel}
\usepackage{microtype}
\usepackage{inconsolata}
\usepackage{graphicx}
\usepackage{booktabs}
\usepackage{multirow}
\usepackage{xcolor}
\usepackage{hyperref}
\usepackage{listings}       
  \usepackage{booktabs}                                    
\usepackage{tikz}
\usetikzlibrary{shapes.geometric, arrows.meta, positioning}
\usepackage{booktabs}
\usepackage{etoolbox}
\AtBeginEnvironment{table}{\nolinenumbers}
\AtBeginEnvironment{table*}{\nolinenumbers}
\AtBeginEnvironment{figure}{\nolinenumbers}
\AtBeginEnvironment{figure*}{\nolinenumbers}

\newcommand{\symbolimg}[2][0.3cm]{%
  \ensuremath{\vcenter{\hbox{\includegraphics[height=#1]{#2}}}}%
}
\newcommand{\openaiicon}{\symbolimg[0.30cm]{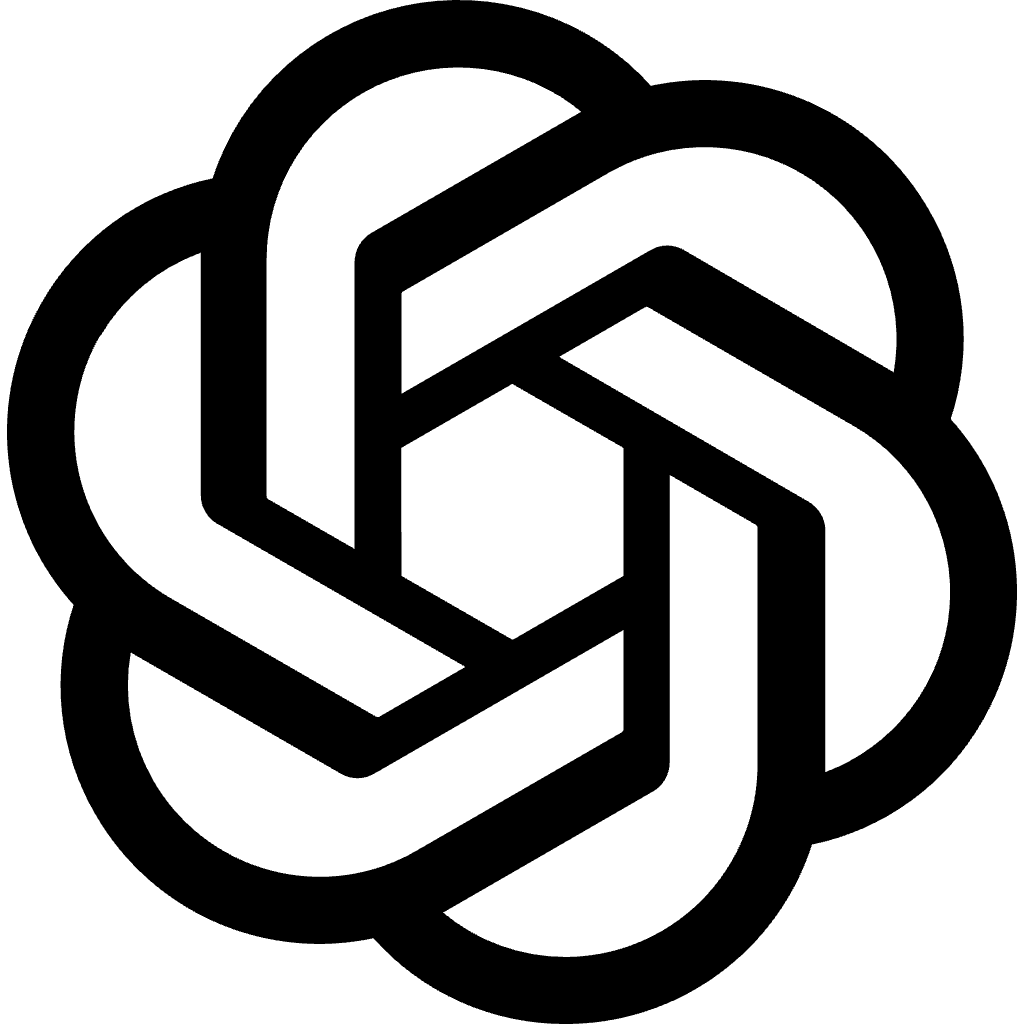}}
\newcommand{\claudeicon}{\symbolimg[0.30cm]{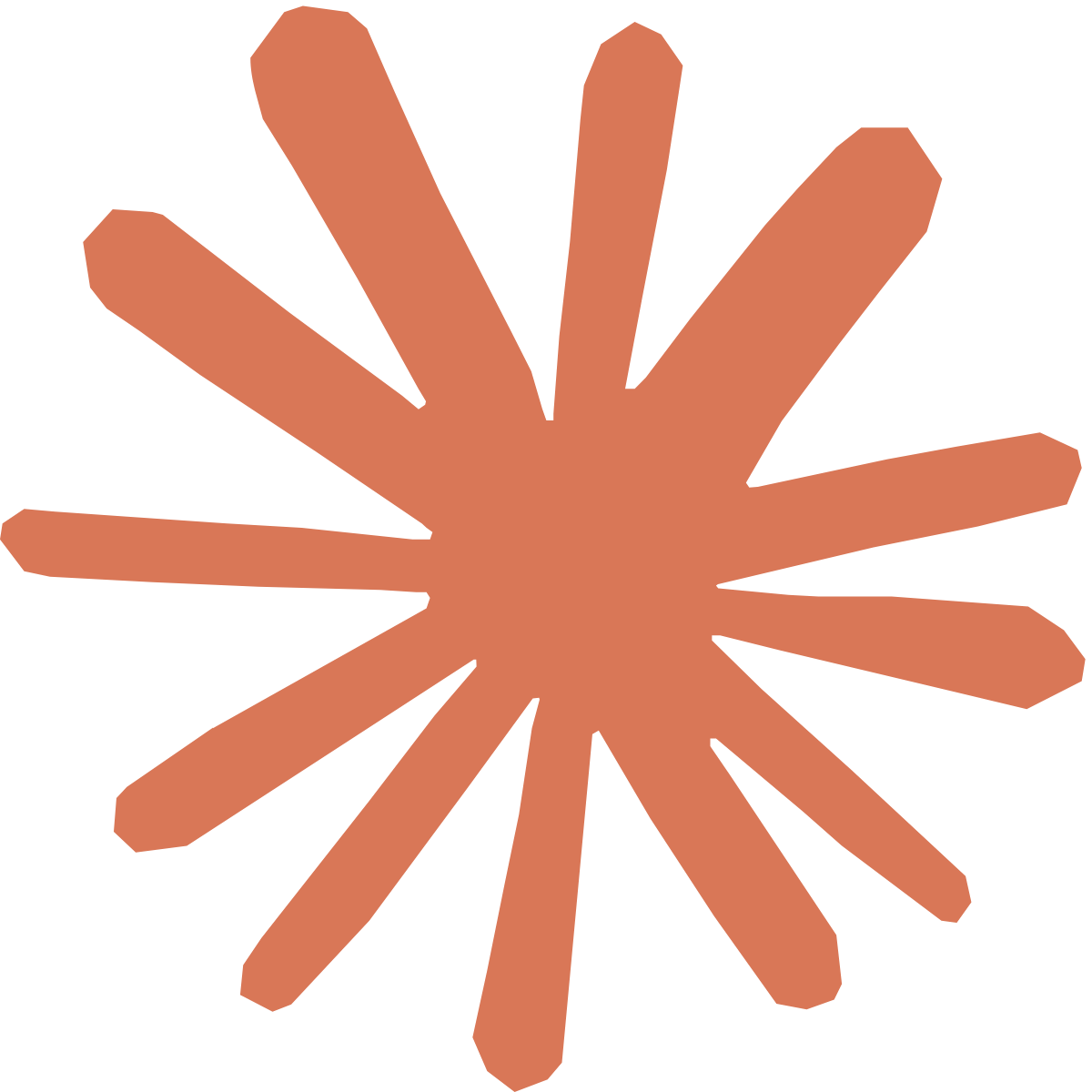}}
\newcommand{\geminiicon}{\symbolimg[0.30cm]{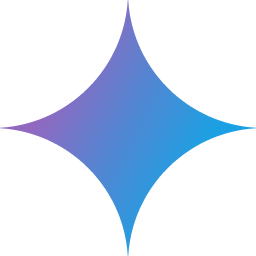}}
\newcommand{\lapaicon}{\symbolimg[0.30cm]{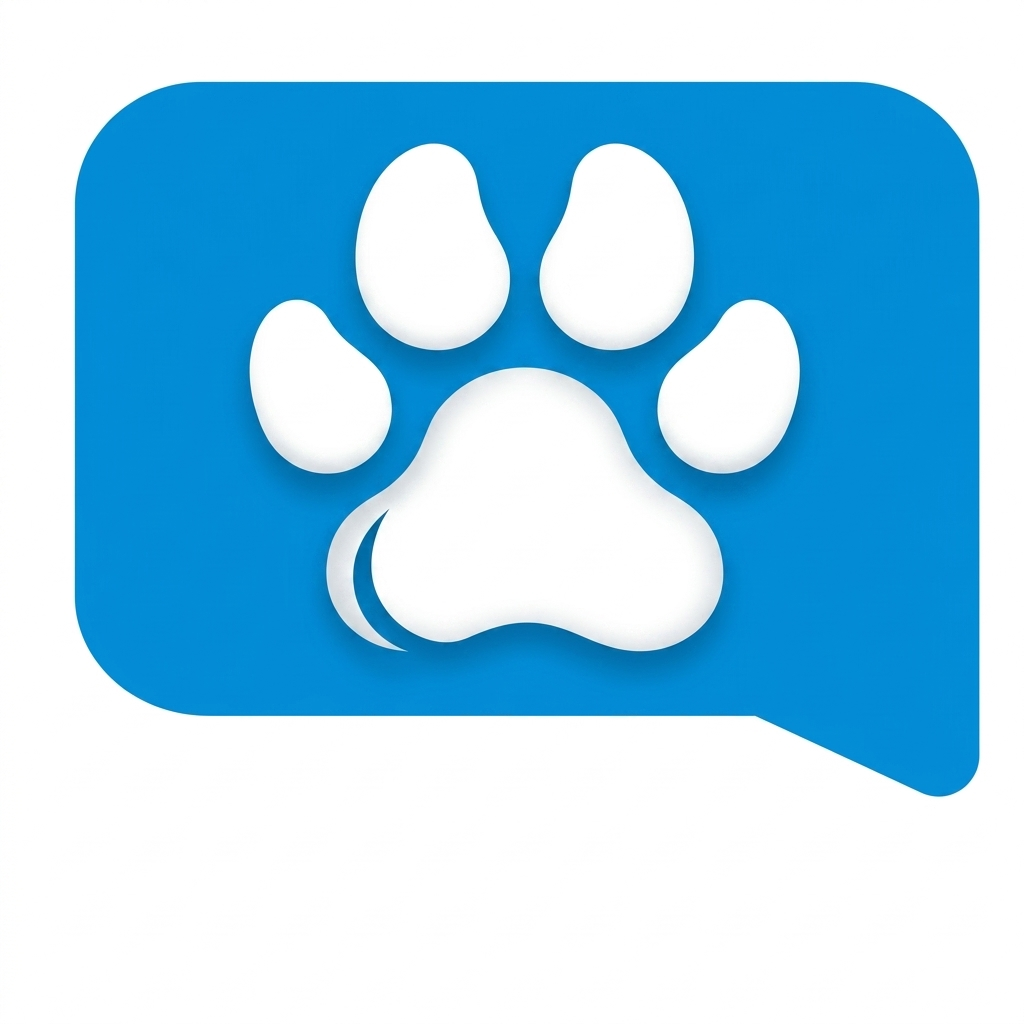}}
\newcommand{\kimiicon}{\symbolimg[0.30cm]{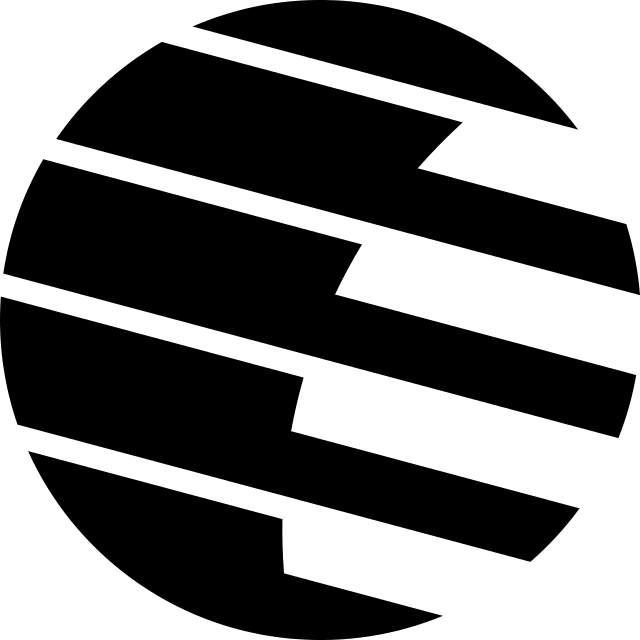}}

\title{How Far Can Prompting Go for Minimal-Edit Ukrainian Grammatical Error Correction?\thanks{This
paper has been accepted for publication at The Fifth Ukrainian Natural Language Processing Conference
UNLP 2026, \url{https://unlp.org.ua/}.}}

\author{
  Kateryna Karpo$^{\upsilon,\sigma}$ \quad Artem Chernodub$^{\zeta}$ \\
  $^{\upsilon}$Ukrainian Catholic University \quad $^{\sigma}$YouScan \quad $^{\zeta}$Zendesk
}

\begin{document}
\maketitle
\begin{abstract}
Fine-tuned Large Language Models (LLMs) dominate in Ukrainian grammatical error correction (GEC), while API-accessed
LLMs remain nearly untested on minimal-edit benchmarks. We evaluate 11 commercial LLMs from four providers and one
open-source Ukrainian model on the UNLP 2023 GEC-only benchmark, comparing zero-shot, few-shot, minimal-edits, and
LLM-assisted prompt optimization strategies. Our best configuration (Gemini~3.1-Pro) reaches $F_{0.5}=69.22$, closing
over 90\% of the gap to fine-tuned SOTA ($F_{0.5}=73.14$). For zero-shot prompts, only Claude models benefit from
Ukrainian instructions. However, the best overall results for all models use Ukrainian minimal-edits prompts, whose
language-specific rules require Ukrainian to express precisely. LLM-assisted prompt optimization on top of minimal-edits
+ few-shot achieves the highest score. Detailed minimal-edits instructions yield the largest gains for punctuation and
case errors but cause the model to abandon several low-frequency categories. Delving into error analysis, we identify
five recurring overcorrection patterns tied to Ukrainian-specific linguistic phenomena. Code, prompts, and outputs are
publicly available.\footnote{Correspondence:
\texttt{a.chernodub@gmail.com}}\footnote{\url{https://github.com/katerynkarpo/gec_unlp_2026}}\footnote{This work was conducted as part of Kateryna Karpo’s M.Sc. thesis at
the Ukrainian Catholic University, Faculty of Applied Sciences.}
\end{abstract}

\section{Introduction and Related Work}

Grammatical error correction (GEC) systems operate under two paradigms~\citep{bryant-etal-2023-gec}. Minimal-edit
correction targets only clear grammatical, spelling, and punctuation errors, preserving the author's wording.
Fluency-oriented correction additionally permits lexical substitutions, syntactic restructuring, and stylistic
improvements. The minimal-edit setting is especially relevant for educational tools, where feedback should pinpoint
errors rather than rewrite learner text, and for writing assistants that must preserve authorial voice.

For English, a high-resource language with decades of GEC research, this distinction is well established. Minimal-edit
evaluation is the standard in shared tasks such as CoNLL-2014~\citep{ng-etal-2014-conll} and
BEA-2019~\citep{bryant-etal-2019-bea}, while JFLEG~\citep{napoles-etal-2017-jfleg} targets fluency.
\citet{staruch-etal-2025-adapting} recently achieved state-of-the-art single-model minimal-edit results on BEA-2019 by
adapting a decoder-only LLM. The MultiGEC-2025 shared task~\citep{masciolini-etal-2025-multigec} extended the two-track
paradigm to twelve European languages, confirming it as a cross-lingual standard.

For Ukrainian, GEC infrastructure has only recently begun to emerge. The UNLP 2023 Shared
Task~\citep{syvokon-romanyshyn-2023-unlp} introduced the first benchmark with two parallel tracks: GEC-only
(minimal-edit) and GEC+Fluency, both evaluated with span-based $F_{0.5}$. Since then, research has shifted toward
fluency~\citep{saini-etal-2024-spivavtor}, with \citet{luhtaru-etal-2024-err} pushing GEC+Fluency SOTA to
$F_{0.5}=74.09$ with a fine-tuned Llama~2 model, surpassing the original winner ($F_{0.5}=68.17$;
\citealp{bondarenko-etal-2023-comparative}). The GEC-only track, however, has seen no new results. Ukrainian was also
included in MultiGEC-2025, where the winning team's fine-tuned Gemma~2 scored $\mathrm{GLEU}=79.55$ on minimal edits vs.\
$\mathrm{GLEU}=68.03$ for a one-shot Llama~3.1 baseline. Most recently, \citet{kovalchuk-etal-2025-omnigec} introduced
silver-standard GEC corpora for multiple languages including Ukrainian and fine-tuned multilingual models on them;
however, their work centers on training data creation and finetuning rather than prompting strategies for API-accessed
LLMs.

To the best of our knowledge, most of Ukrainian GEC systems available to date rely on fine-tuned models that require
dedicated GPU infrastructure. Commercial API-accessed LLMs offer a lightweight alternative, yet remain nearly untested
for Ukrainian minimal-edit GEC. The only published result is from \citet{katinskaia-yangarber-2024-gpt}, who evaluated
GPT-3.5 (specifically, \texttt{gpt-3.5-turbo-0613}) in a zero-shot setting on the UNLP 2023 GEC-only test set and
obtained $F_{0.5}=27.4$, far below the fine-tuned
SOTA of $73.14$.

This paper is the first to test newer API-accessed models on this benchmark and to explore whether better prompting
strategies can close the gap with fine-tuned systems.

\section{Experimental Setup}

\paragraph{Data.} We use the GEC-only track of the UNLP 2023 Shared Task~\citep{syvokon-romanyshyn-2023-unlp}, which is
built on the UA-GEC corpus. We adopt UA-GEC's own train and valid splits as our training and development sets (31,038
and 1,422 sentences) and report all final numbers on the UNLP 2023 test set (1,274 sentences), whose gold annotations
are held out from participants and never inspected during prompt development. We use the train and development sets
for prompt development (few-shot exemplar selection and prompt engineering) and report only on the test set.

\paragraph{Models.} We evaluate commercial, API-accessed LLMs from four providers and one open-source Ukrainian model,
Lapa~v0.1.2~\citep{paniv-etal-2025-lapa}.\footnote{Provider documentation: OpenAI \url{https://platform.openai.com};
Anthropic \url{https://docs.anthropic.com}; Google \url{https://ai.google.dev}; Moonshot
\url{https://platform.moonshot.ai}.} We report exact snapshot identifiers for reproducibility. From OpenAI, we use
GPT-4.1 (\texttt{gpt-4.1-2025-04-14}), GPT-4.1-mini (\texttt{gpt-4.1-mini-2025-04-14}), GPT-5.1
(\texttt{gpt-5.1-2025-11-13}), GPT-5.2 (\texttt{gpt-5.2-2025-12-11}), and GPT-5.4 (\texttt{gpt-5.4-2026-03-05}). From
Moonshot, we use Kimi-K2 (\texttt{kimi-k2-0905-preview}; \texttt{0905} denotes a dated preview build). The Google and
Anthropic APIs do not expose dated snapshot identifiers; we use Gemini~3-Flash (\texttt{gemini-3-flash-preview}),
Gemini~3-Pro (\texttt{gemini-3-pro-preview}), Gemini~3.1-Pro (\texttt{gemini-3.1-pro-preview}), Claude Sonnet~4.6
(\texttt{claude-sonnet-4.6}), and Claude Opus~4.6
(\texttt{claude-opus-4.6}).\footnote{Inference-time parameters differ: GPT-4.1 and Kimi use temperature/top-$p$; Claude and Gemini use either
temperature or a reasoning effort budget; GPT-5.x uses an effort level (\texttt{low}/\texttt{medium}/\texttt{high}). We
set temperature~0 where available, default effort for Claude and Gemini, and \texttt{medium} for GPT-5.x.}

\subsection{Research Questions}

We address the following four research questions:

\paragraph{RQ1: What is the zero-shot minimal-edit GEC performance of current LLMs on Ukrainian relative to fine-tuned
SOTA, and how sensitive is it to prompt language?}
We systematically compare 2025--2026 commercial LLMs on the UNLP 2023 minimal-edit GEC benchmark against the fine-tuned
SOTA of $F_{0.5}=73.14$, and test both English and Ukrainian prompt variants to assess whether instruction language
affects correction quality for a morphologically rich, low-resource language. To the best of our knowledge, the only
published LLM baseline for Ukrainian GEC is the GPT-3.5 zero-shot result (English) from
\citet{katinskaia-yangarber-2024-gpt}, which we include for reference.

\paragraph{RQ2: Can prompting strategies reduce overcorrection compared to zero-shot baselines?}
We evaluate how each of the four prompting strategies affects the precision--recall trade-off: (1)~zero-shot,
(2)~few-shot, (3)~minimal-edits + zero-shot, and (4)~minimal-edits + few-shot.

\paragraph{RQ3: Can LLM-assisted prompt optimization improve over manually crafted prompts?}
We apply an LLM-assisted prompt optimization pipeline built on Claude Code skills, an agentic system that iteratively
generates, evaluates, and refines GEC prompts using the full evaluation loop as feedback.

\paragraph{RQ4: Where do minimal-edits instructions help and where do they fail?}
We compare per-error-type performance between a standard zero-shot prompt and our best optimized prompt using ERRANT
category breakdowns, identifying which error types benefit most from detailed minimal-edits instructions and which
remain resistant to prompt-based improvement.

\paragraph{Prompting strategies.} We compare four manually engineered prompting configurations that vary in prompt
detail (general vs.\ minimal-edits) and use of examples (zero-shot vs.\ few-shot).

\begin{enumerate}
\item \textbf{Zero-shot} (\ref{prompt:zero-shot}): a general system prompt that instructs the model to correct
grammatical and spelling errors and return the original sentence if no errors are found. No examples are provided.

\item \textbf{Few-shot} (\ref{prompt:few-shot}): the zero-shot prompt augmented with source--target correction pairs
from the training set, covering spelling, punctuation, and morphological errors as well as already-correct sentences.

\item \textbf{Minimal-edits + zero-shot} (\ref{prompt:minimal-edits-zero-shot}): a detailed system prompt enumerating
which error types to correct, Ukrainian-specific conventions (e.g., dash vs.\ hyphen in dialogue,
\foreignlanguage{ukrainian}{\emph{у/в}} `u/v' alternation, vocative case in forms of address, etc.), and categories of
changes to avoid. No examples are provided.

\item \textbf{Minimal-edits + few-shot} (\ref{prompt:minimal-edits-few-shot}): combines the detailed minimal-edits
system prompt with few-shot correction examples from the training set, providing both rule-based guidance and concrete
demonstrations.

\end{enumerate}

\noindent Strategies~(1)--(4) are tested with both English (EN) and Ukrainian (UA) prompt text to address RQ1. For
subsequent experiments (RQ2--RQ4), we use EN for zero-shot and few-shot prompts (where it performs best for most models;
see RQ1) and UA for minimal-edits variants. This was an intentional design choice: the minimal-edits rules reference
specific Ukrainian word forms, morphological categories, and language-specific conventions (e.g., vocative case
paradigms, euphonic preposition alternation) that cannot be adequately expressed in English. Because the prompt language
and strategy are tied together in this comparison, we treat them as a single design decision. To test whether an LLM can
improve over these handcrafted prompts, we also apply LLM-assisted prompt optimization, inspired by automatic prompt
optimization methods \cite[see][for a survey]{ramnath-etal-2025-systematic}, on top of the best minimal-edits + few-shot
prompt (Appendix~\ref{prompt:apo}; RQ3).

\paragraph{Evaluation.} We use the official UNLP 2023 evaluation pipeline, which computes span-level Precision~(P),
Recall~(R), and $F_{0.5}$ using a Ukrainian adaptation of ERRANT. Per-error-type scores are extracted from the ERRANT
alignment for the per-error-type analysis (RQ4).

\section{Prompt Design}
\label{sec:prompt-design}

\paragraph{Zero-shot.}
We use a single-sentence system prompt (Appendix~\ref{prompt:zero-shot}), adapted from \citet{loem-etal-2023-exploring}:

\begin{quote}\small
\texttt{Reply with a corrected version of the sentence with all grammatical and spelling errors fixed. If there are no
errors, reply with a copy of the original sentence. Input sentence: \{sentence\}. Corrected sentence:}
\end{quote}

\noindent The Ukrainian version is its direct translation:

\begin{quote}\small
\foreignlanguage{ukrainian}{\texttt{Надай виправлену версію речення з виправленими всіма граматичними та орфографічними
помилками. Якщо помилок немає, надай копію оригінального речення. Вхідне речення: \{sentence\}. Виправлене речення:}}
\end{quote}

\paragraph{Few-shot.}
The few-shot prompt extends the zero-shot instruction with source--target correction pairs from the training set
(Appendix~\ref{prompt:few-shot}), e.g.:

\begin{quote}\small
\texttt{[Same header as zero-shot prompt]}\\[2pt]
\texttt{Input:} \foreignlanguage{ukrainian}{\texttt{Так само потерпає Україна і сьогодні від того що насправді
талановитим людям заважають працювати...}}\\
\texttt{Output:} \foreignlanguage{ukrainian}{\texttt{Так само потерпає Україна і сьогодні від того, що насправді
талановитим людям заважають працювати...}}\\[2pt]
\textit{(Input: `Ukraine suffers the same today from the fact that truly talented people are prevented from
working...'\\
Output: `Ukraine suffers the same today from the fact\textbf{,} that truly talented people are prevented from
working...')}
\end{quote}

Exemplars are drawn from the UA-GEC training split because it is the only publicly available Ukrainian GEC corpus
at the required scale and annotation quality. Since this split is public, it may have been seen by commercial LLMs
during pretraining; drawing exemplars from an independent Ukrainian GEC corpus would be a cleaner control, but no
comparable dataset currently exists. We therefore treat our numbers as establishing prompting baselines on this
benchmark and revisit this risk in the Limitations section.

\paragraph{Minimal-edits.}
The zero-shot and few-shot prompts give only a generic correction instruction (``fix all grammatical and spelling
errors''), which provides no guidance on correction scope. In practice, this leads LLMs to overcorrect: rephrasing
sentences, substituting synonyms, or ``improving'' stylistically acceptable constructions. Since the ERRANT-based
$F_{0.5}$ metric penalizes unnecessary edits, such overcorrection directly hurts precision.

The minimal-edits prompt addresses this with a two-part structure (Appendix~\ref{prompt:minimal-edits-zero-shot}). The
first part explicitly declares the minimal-edit constraint: ``correct only clear-cut errors while preserving the
original wording``. The second part provides a specific taxonomy of 16 Ukrainian GEC error categories (spelling,
punctuation, case, gender, number, aspect, tense, etc.), followed by language-specific conventions (e.g.,
\foreignlanguage{ukrainian}{у/в} `u/v' alternation before consonants/vowels, em-dash in dialogue) and strict rules on
what not to change (no synonym substitution, no quote style normalization, no changes when in doubt):

\begin{quote}\small
\texttt{...}\\[2pt]
\foreignlanguage{ukrainian}{\texttt{Виправляй ЛИШЕ такі типи помилок:}}\\
\textit{(`Fix ONLY the following types of errors:')}\\
\foreignlanguage{ukrainian}{\texttt{1. Орфографія: явні орфографічні помилки}}\\
\textit{(`1. Spelling: obvious spelling errors')}\\
\foreignlanguage{ukrainian}{\texttt{2. Пунктуація: пропущені або зайві коми, крапки...}}\\
\textit{(`2. Punctuation: missing or extra commas, periods...')}\\
\foreignlanguage{ukrainian}{\texttt{3. G/Case: некоректне вживання відмінкової форми}}\\
\textit{(`3. G/Case: incorrect use of case form')}\\
\texttt{...}
\end{quote}

\noindent The minimal-edits + few-shot variant (Appendix~\ref{prompt:minimal-edits-few-shot}) combines this detailed
system prompt with few-shot examples.

\paragraph{LLM-assisted prompt optimization.}
Inspired by automatic prompt optimization methods \citep{ramnath-etal-2025-systematic}, we develop a semi-automatic
approach in which an LLM proposes prompt edits but a human reviews and accepts them. Our method borrows ideas from
several automatic prompt optimization papers: like ProTeGi~\citep{pryzant-etal-2023-protegi}, we use LLM-generated
``textual gradients'' derived from error analysis to guide prompt edits; following
PromptAgent~\citep{wang-etal-2024-promptagent}, we cluster prediction--reference mismatches into recurring linguistic
patterns (e.g., ``unnecessary dash normalization'', ``missed comma before subordinate conjunction'') to produce
domain-expert-style prompt sections; and as in OPRO~\citep{yang-etal-2024-opro}, we maintain an optimization history of
previous candidates and their scores to inform each iteration.

\paragraph{LLM-assisted prompt optimization design.} We implemented this pipeline as a Claude Code skill powered by
Claude Opus~4.6, which acts as both error analyst and prompt engineer. Starting from the best manual prompt (usually
minimal-edits + few-shot), the agent iteratively:
(1)~evaluates the candidate on the validation set, recording span-level TP/FP/FN;
(2)~clusters mismatches into linguistic patterns ranked by frequency;
(3)~modifies the prompt via rule insertion (an explicit prohibition in the ``do not change'' section) or example
insertion (a targeted input--output pair, including ``no-change'' examples);
(4)~accepts the change only if $F_{0.5}$ improves, otherwise reverts.
The cycle repeats until gains plateau.

\paragraph{Optimization setup.} Due to cost and time constraints, we could not run the optimization loop separately for
every model. Instead, we selected the best-performing manual prompt, minimal-edits + few-shot
(\ref{prompt:minimal-edits-few-shot}), and optimized it in two rounds: first on GPT-4.1-mini, producing minimal-edits +
few-shot + optimized-v1 (\ref{prompt:apo A002}), and then starting from that result on Gemini~3-Flash, producing
minimal-edits + few-shot + optimized-v2 (\ref{prompt:apo}). We then transferred these prompts to the remaining models
without further tuning: GPT and Claude models are evaluated with optimized-v1, Gemini models with optimized-v2
(Table~\ref{tab:rq3-apo}). We acknowledge that per-model optimization would give a more complete picture; we report
these preliminary results as a useful reference point.

\begin{figure}[t!]
  \centering
  \includegraphics[width=\columnwidth]{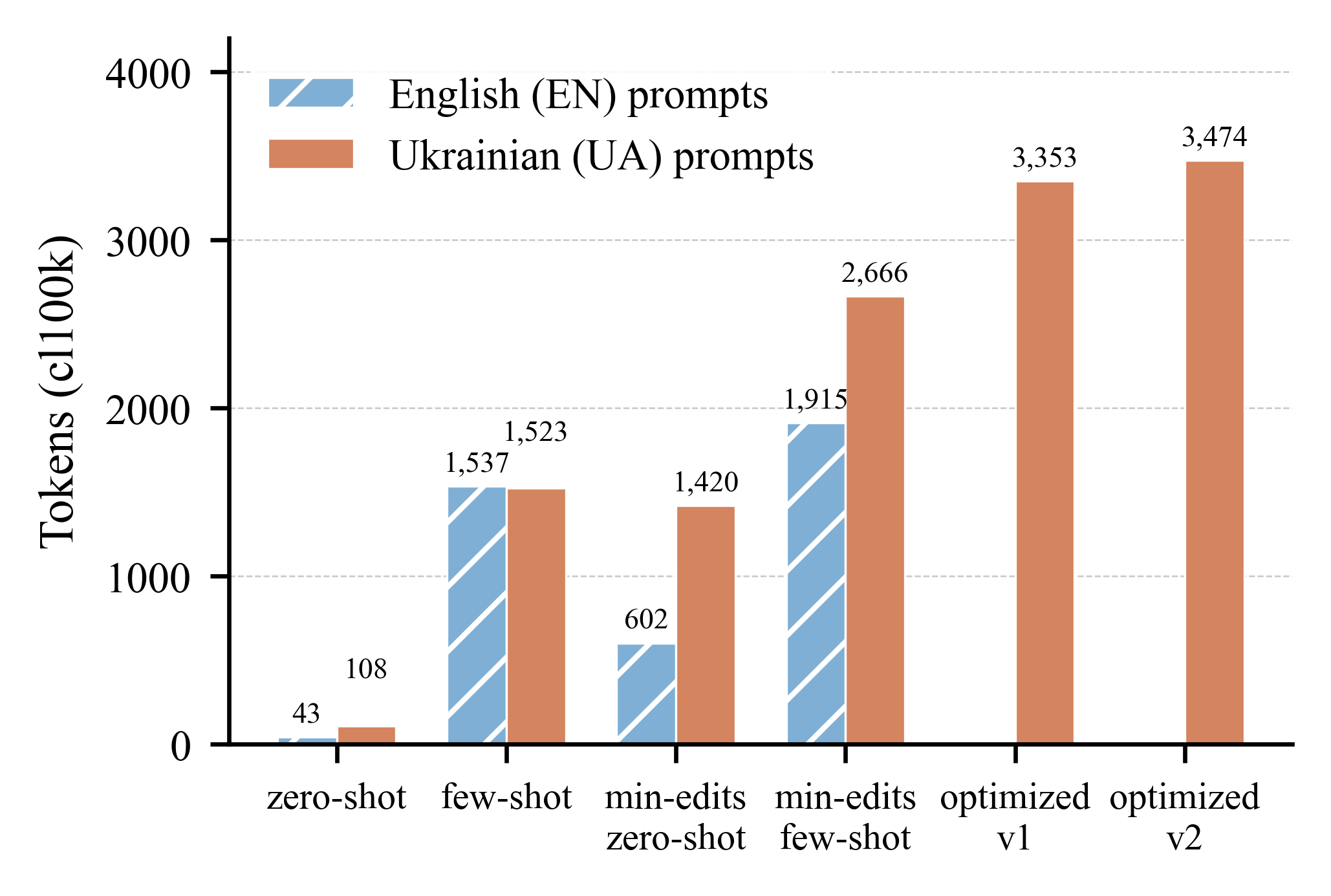}
  \caption[Prompt length across prompting strategies]{Prompt length (in tokens, cl100k\_base tokenizer\protect\footnotemark) across prompting strategies for English (EN) and Ukrainian (UA) prompts. Optimized variants are available in UA only. The minimal-edits + few-shot + optimized-v2 prompt (\ref{prompt:apo}) is ${\sim}32{\times}$ longer than the zero-shot UA baseline (\ref{prompt:zero-shot-ua}; 3{,}474 vs.\ 108 tokens).}
  \label{fig:prompt-tokens}
\end{figure}

\paragraph{Prompt length.} Figure~\ref{fig:prompt-tokens} shows how prompt length grows across strategies, from 43
tokens for zero-shot (EN, \ref{prompt:zero-shot-en}) to 3{,}474 tokens for minimal-edits + few-shot + optimized-v2 (UA,
\ref{prompt:apo}).

% ===== FIGURE: Prompt Token Counts =====
\footnotetext{\url{https://github.com/openai/tiktoken}}

\section{Experimental Results}

\paragraph{RQ1: Zero-shot performance and prompt language (Table~\ref{tab:rq1-language}).}

% ===== TABLE 1: RQ1 — Prompt Language =====
\begin{table}[t!]
\centering
\footnotesize
\setlength{\tabcolsep}{3pt}
\begin{tabular}{llrrrc@{}}
\toprule
\textbf{Model} & \textbf{Lang.} & \textbf{Prec.} & \textbf{Rec.} & $\mathbf{F_{0.5}}$ & \textbf{UA?} \\
\midrule
\multicolumn{5}{l}{\textit{Baseline: fine-tuned}} \\
\quad mBART50-large$^\dagger$ & -- & 78.52 & 50.60 & 70.71 & \\
\quad mT5-large$^\ddagger$ & -- & \textbf{76.81} & \textbf{61.39}  & \textbf{73.14} & \\
\midrule\midrule
\multicolumn{5}{l}{\textit{Baseline: LLM zero-shot}} \\
\quad \openaiicon~GPT-3.5$^*$ & EN & 25.80 & 36.20 & 27.40 & \\
\midrule
\multicolumn{5}{l}{\textit{API-accessed (zero-shot)}} \\
\quad \openaiicon~GPT-4.1                   & EN & 37.17 & 55.54 & 39.80 & \\
\quad \openaiicon~GPT-4.1                   & UA & 30.24 & 58.41 & 33.47 & \\
\midrule
\quad \openaiicon~GPT-4.1-mini              & EN & 39.06 & 51.92 & 41.09 & \\
\quad \openaiicon~GPT-4.1-mini              & UA & 36.73 & 53.03 & 39.13 & \\
\midrule
\quad \openaiicon~GPT-5.1 (medium)          & EN & 36.06 & 60.15 & 39.20 & \\
\quad \openaiicon~GPT-5.1 (medium)          & UA & 32.35 & 62.61 & 35.82 & \\
\midrule
\quad \openaiicon~GPT-5.2 (medium)          & EN & 34.04 & 66.31 & 37.71 & \\
\quad \openaiicon~GPT-5.2 (medium)          & UA & 29.89 & 66.90 & 33.60 & \\
\midrule
\quad \openaiicon~GPT-5.4 (medium)          & EN & 36.87 & 62.96 & 40.20 & \\
\quad \openaiicon~GPT-5.4 (medium)          & UA & 32.73 & 65.35 & 36.36 & \\
\midrule
\quad \claudeicon~Claude Sonnet~4.6         & EN & 41.79 & 45.83 & 42.54 & \\
\quad \claudeicon~Claude Sonnet~4.6         & UA & 42.70 & 47.76 & 43.63 & $+$ \\
\midrule
\quad \claudeicon~Claude Opus~4.6           & EN & 47.60 & 46.20 & 47.30 & \\
\quad \claudeicon~Claude Opus~4.6           & UA & \textbf{49.20} & 51.60 & \textbf{49.70} & $+$ \\
\midrule
\quad \geminiicon~Gemini~3-Flash            & EN & 39.09 & 64.85 & 42.46 & \\
\quad \geminiicon~Gemini~3-Flash            & UA & 35.91 & \textbf{67.38} & 39.61 & \\
\midrule
\quad \geminiicon~Gemini~3-Pro              & EN & 37.89 & 60.54 & 40.96 & \\
\quad \geminiicon~Gemini~3-Pro              & UA & 36.30 & 63.58 & 39.71 & \\
\midrule
\quad \geminiicon~Gemini~3.1-Pro            & EN & 41.50 & 58.03 & 44.01 & \\
\quad \geminiicon~Gemini~3.1-Pro            & UA & 39.16 & 62.01 & 42.28 & \\
\midrule
\quad \kimiicon~Kimi-K2                     & EN & 34.24 & 52.74 & 36.82 & \\
\quad \kimiicon~Kimi-K2                     & UA & 29.03 & 60.11 & 32.38 & \\
\midrule
\multicolumn{5}{l}{\textit{Open-source (zero-shot)}} \\
\quad \lapaicon~Lapa~v0.1.2$^\S$ & EN & 24.24 & 23.52 & 24.09 & \\
\quad \lapaicon~Lapa~v0.1.2$^\S$ & UA & 28.57 & 32.38 & 29.26 & $+$ \\
\bottomrule
\end{tabular}
\caption{RQ1: Zero-shot GEC performance on the UNLP 2023 test set.
Lang.\ denotes the language of the zero-shot instructions in the system prompt: Ukrainian (UA) or
English (EN). Bold marks the best result per column among API-accessed models.
UA?: $+$ indicates that the UA prompt yields a higher $F_{0.5}$ than the EN prompt for the same model.
$^\dagger$\citet{syvokon-romanyshyn-2023-unlp}; $^\ddagger$\citet{gomez-etal-2023-low};
$^*$\citet{katinskaia-yangarber-2024-gpt}, EN; $^\S$\citet{paniv-etal-2025-lapa}, Ukrainian open-source LLM.}
\label{tab:rq1-language}
\end{table}
We start with zero-shot prompts, the simplest and most widely used setup for LLM-based GEC, and test whether prompting
in Ukrainian rather than English improves results.

\paragraph{Zero-shot performance.}
All current models dramatically outperform the GPT-3.5 baseline of $F_{0.5}=27.4$ reported by
\citet{katinskaia-yangarber-2024-gpt}.
The best zero-shot system, Claude Opus~4.6 with a UA prompt, reaches $F_{0.5}=49.70$, nearly doubling the GPT-3.5 score.
Notably, GPT-5.x reasoning models do not outperform the older GPT-4.1-mini ($F_{0.5}=41.09$), despite their stronger
general benchmarks. We attribute this to overcorrection: reasoning-optimized models tend to over-interpret the
correction task, producing more extensive rewrites that ERRANT penalizes as false positives.

Even the best zero-shot result remains 23.4 $F_{0.5}$ points below the fine-tuned SOTA of 73.14. Comparing the best
zero-shot system (Claude Opus~4.6~UA: P$=$49.20, R$=$51.60) against the fine-tuned reference (P$=$76.81, R$=$61.39), the
gap is primarily driven by precision (27.6-point difference) rather than recall (9.8-point difference). Without
task-specific fine-tuning, zero-shot models lack the calibration to suppress spurious corrections.

Models differ in their precision--recall profiles. Gemini variants are high-recall correctors (R$=$60--67\%) with low
precision (P$=$36--42\%), flagging many candidates, many of which are spurious. Claude models show a more balanced
profile with precision and recall within 5 points of each other, which is favorable under $F_{0.5}$'s precision
weighting. GPT-5.x models lean toward high recall and low precision, similar to Gemini.

\textbf{Prompt language.}
Surprisingly, Claude is the only family where UA prompts improve performance: Claude Opus~4.6 gains 2.4~points (47.30
$\to$ 49.70) and Claude Sonnet~4.6 gains 1.1~points (42.54 $\to$ 43.63), with both precision and recall improving
simultaneously.
For all other models, UA prompts degrade $F_{0.5}$ by 1--6 points, consistently trading precision for recall and
amplifying overcorrection.
For reference, we also include Lapa~v0.1.2\footnote{Lapa \citep{paniv-etal-2025-lapa} is an open-source Ukrainian LLM
fine-tuned with GEC-style prompts different from ours, and the only non-API-accessed model in our evaluation. We include
it as a reference point for open-weight Ukrainian-centric models, though a full comparison with open-source alternatives
is beyond the scope of this work.}, an open-source Ukrainian LLM, which scores $F_{0.5}=29.26$ with a UA prompt, above
GPT-3.5 but well below the API-accessed models.
% ===== TABLE 2: RQ2 — Prompting Strategies =====
\begin{table*}[t!]
\centering
\small
\begin{tabular}{lllrrr@{}}
\toprule
\textbf{Model} & \textbf{Prompting Strategy} & \textbf{Lang.} & \textbf{Prec.} & \textbf{Rec.} & $\mathbf{F_{0.5}}$ \\
\midrule
\multicolumn{5}{l}{\textit{Baseline: fine-tuned SOTA}} \\
\quad mT5-large$^\ddagger$ & -- & --    & 76.81 & 61.39  & 73.14 \\
\midrule\midrule
\openaiicon~GPT-4.1-mini  & zero-shot (\ref{prompt:zero-shot-en})                  & EN & 39.06 & 51.92 & 41.09 \\
              & few-shot (\ref{prompt:few-shot-en})                   & EN & 44.75 & 59.73 & 47.11 \\
              & minimal-edits + zero-shot (\ref{prompt:minimal-edits-zero-shot-ua})    & UA & 46.66 & 51.52 & 47.56 \\
              & minimal-edits + few-shot (\ref{prompt:minimal-edits-few-shot-ua})     & UA & 47.16 & 51.48 & 47.97 \\
\midrule
\openaiicon~GPT-5.4       & zero-shot (\ref{prompt:zero-shot-en})                  & EN & 36.87 & 62.96 & 40.20 \\
              & few-shot (\ref{prompt:few-shot-en})                   & EN & 46.24 & 66.67 & 49.26 \\
              & minimal-edits + zero-shot (\ref{prompt:minimal-edits-zero-shot-ua})    & UA & 57.05 & 63.18 & 58.18 \\
              & minimal-edits + few-shot (\ref{prompt:minimal-edits-few-shot-ua})     & UA & 60.02 & 58.40 & 59.69 \\
\midrule
\claudeicon~Claude Sonnet~4.6 & zero-shot (\ref{prompt:zero-shot-en})              & EN & 41.79 & 45.83 & 42.54 \\
              & few-shot (\ref{prompt:few-shot-en})                   & EN & 52.52 & 56.26 & 53.23 \\
              & minimal-edits + zero-shot (\ref{prompt:minimal-edits-zero-shot-ua})    & UA & 62.44 & 48.55 & 59.06 \\
              & minimal-edits + few-shot (\ref{prompt:minimal-edits-few-shot-ua})     & UA & 61.96 & 49.10 & 58.88 \\
\midrule
\claudeicon~Claude Opus~4.6 & zero-shot (\ref{prompt:zero-shot-en})                & EN & 47.60 & 46.20 & 47.30 \\
              & few-shot (\ref{prompt:few-shot-en})                   & EN & 56.83 & 55.93 & 56.65 \\
              & minimal-edits + zero-shot (\ref{prompt:minimal-edits-zero-shot-ua})    & UA & 67.63 & 47.08 & 62.20 \\
              & minimal-edits + few-shot (\ref{prompt:minimal-edits-few-shot-ua})     & UA & \textbf{68.54} & 49.75 &
              63.73 \\
\midrule
\geminiicon~Gemini~3-Flash & zero-shot (\ref{prompt:zero-shot-en})                 & EN & 39.09 & 64.85 & 42.46 \\
              & few-shot (\ref{prompt:few-shot-en})                   & EN & 48.48 & \textbf{72.01} & 51.87 \\
              & minimal-edits + zero-shot (\ref{prompt:minimal-edits-zero-shot-ua})    & UA & 53.29 & 66.40 & 55.48 \\
              & minimal-edits + few-shot (\ref{prompt:minimal-edits-few-shot-ua})     & UA & 60.28 & 66.18 & 61.38 \\
\midrule
\geminiicon~Gemini~3.1-Pro & zero-shot (\ref{prompt:zero-shot-en})                 & EN & 41.50 & 58.03 & 44.01 \\
              & few-shot (\ref{prompt:few-shot-en})                   & EN & 54.92 & 66.25 & 56.86 \\
              & minimal-edits + zero-shot (\ref{prompt:minimal-edits-zero-shot-ua})    & UA & 60.49 & 65.24 & 61.38 \\
              & minimal-edits + few-shot (\ref{prompt:minimal-edits-few-shot-ua})     & UA & 63.76 & 63.35 &
              \textbf{63.68} \\
\bottomrule
\end{tabular}
\caption{RQ2: Effect of prompting strategies on the UNLP 2023 test set. Zero-shot and few-shot use EN prompts;
minimal-edits variants use UA prompts (see Section~\ref{sec:prompt-design} for rationale).
Lang.\ denotes the language of the zero-shot instructions in the system prompt: Ukrainian (UA) or English (EN).
For each model, we report the best configuration per strategy. Bold marks the best result per column among API-accessed
models.
$^\ddagger$\citet{gomez-etal-2023-low}.}
\label{tab:rq2-strategies}
\end{table*}

\paragraph{RQ2: Prompting strategies (Table~\ref{tab:rq2-strategies}).}
We select the six best-performing models from RQ1 (one to two per provider, excluding lower-scoring variants) and test
whether few-shot examples and minimal-edits constraints can reduce the overcorrection observed in RQ1.

\paragraph{Best results and gap to SOTA.}
Combining few-shot examples with minimal-edits instructions, minimal-edits + few-shot yields the best or near-best
$F_{0.5}$ for every model. Claude Opus~4.6 ($F_{0.5}=63.73$) and Gemini~3.1-Pro (63.68) effectively tie despite
different zero-shot starting points. GPT-5.4 shows the largest absolute gain ($+$19.5 points), recovering from the
weakest zero-shot result to a competitive 59.69.

However, even the best prompted result falls 9.4~points below the fine-tuned SOTA of 73.14, with the gap concentrated in
precision (P$=$76.81 vs.\ 68.54). The minimal-edits instruction suppresses the most egregious false positives, but a
long tail of borderline corrections remains that likely requires task-specific fine-tuning.

\paragraph{Few-shot gains.}
Adding few-shot examples to the zero-shot prompt produces moderate but reliable gains of $+$6--13 $F_{0.5}$ points,
driven by improvements in both precision and recall. The gains are largest for Gemini~3.1-Pro ($+$12.85) and smallest
for GPT-4.1-mini ($+$6.02).

\paragraph{Minimal-edits instructions matter most.}
The minimal-edits constraint has a larger effect than few-shot examples. Switching from a generic EN prompt to a UA
minimal-edits instruction, even without few-shot examples, already matches or exceeds few-shot-only performance for five
out of six models. The most striking case is GPT-5.4: minimal-edits + zero-shot alone yields $F_{0.5}=58.18$, a full
9~points above its few-shot score of 49.26. The mechanism is a sharp precision increase ($+$8--21 points across models)
with modest recall change, meaning the constraint reduces unnecessary edits without hurting the model's ability to catch
real errors.

\paragraph{Overall trends.}
Across all six models, $F_{0.5}$ improves consistently along the progression: zero-shot $<$ few-shot $<$ minimal-edits +
zero-shot $\leq$ minimal-edits + few-shot, with one notable exception: Claude Sonnet~4.6 peaks at minimal-edits +
zero-shot (59.06) and slightly drops with the addition of few-shot examples (58.88). As discussed in
Section~\ref{sec:prompt-design}, the minimal-edits prompts are written in Ukrainian by design, so we cannot fully
separate the effect of prompt language from the effect of the prompting strategy itself.

% ===== TABLE 3: RQ3 — APO =====
\begin{table*}[t!]
\centering
\small
\begin{tabular}{lllrrr@{}}
\toprule
\textbf{Model} & \textbf{Prompting Strategy} & \textbf{Lang.} & \textbf{Prec.} & \textbf{Rec.} & $\mathbf{F_{0.5}}$ \\
\midrule
\multicolumn{5}{l}{\textit{Baseline: fine-tuned SOTA}} \\
\quad mT5-large$^\ddagger$ & -- & -- & 76.81 & 61.39 & 73.14 \\
\midrule\midrule
\openaiicon~GPT-4.1-mini       & minimal-edits + few-shot (\ref{prompt:minimal-edits-few-shot-ua})  & UA & 47.16 & 51.48 & 47.97 \\
                    & minimal-edits + few-shot + optimized-v1 (\ref{prompt:apo A002})  & UA & 55.75 & 51.49 & 54.84 \\
\midrule
\openaiicon~GPT-5.4            & minimal-edits + few-shot (\ref{prompt:minimal-edits-few-shot-ua})  & UA & 60.02 & 58.40 & 59.69 \\
                    & minimal-edits + few-shot + optimized-v1 (\ref{prompt:apo A002})  & UA & 63.94 & 51.75 & 61.07 \\
\midrule
\claudeicon~Claude Sonnet~4.6  & minimal-edits + zero-shot\footnotemark{} (\ref{prompt:minimal-edits-zero-shot-ua}) & UA & 62.44 &
48.55 & 59.06 \\
                    & minimal-edits + few-shot + optimized-v1 (\ref{prompt:apo A002})  & UA & 66.73 & 36.58 & 57.29 \\
\midrule
\claudeicon~Claude Opus~4.6    & minimal-edits + few-shot (\ref{prompt:minimal-edits-few-shot-ua})  & UA & 68.54 & 49.75 & 63.73 \\
                    & minimal-edits + few-shot + optimized-v1 (\ref{prompt:apo A002})  & UA & 66.54 & 38.27 & 57.98 \\
\midrule
\geminiicon~Gemini~3-Flash     & minimal-edits + few-shot (\ref{prompt:minimal-edits-few-shot-ua})  & UA & 60.28 & \textbf{66.18} &
61.38 \\
                    & minimal-edits + few-shot + optimized-v2 (\ref{prompt:apo}) & UA & 69.98 & 62.87 & 68.43 \\
\midrule
\geminiicon~Gemini~3.1-Pro     & minimal-edits + few-shot (\ref{prompt:minimal-edits-few-shot-ua})  & UA & 63.76 & 63.35 & 63.68 \\
                    & minimal-edits + few-shot + optimized-v2 (\ref{prompt:apo}) & UA & \textbf{70.77} & 63.63 &
                    \textbf{69.22} \\
\bottomrule
\end{tabular}
\caption{RQ3: Effect of LLM-assisted prompt optimization, evaluated on the UNLP 2023 test set. Optimized-v1 was
tuned on GPT-4.1-mini, optimized-v2 on Gemini~3-Flash; both were then transferred to other models.
Lang.\ denotes the language of the zero-shot instructions in the system prompt: Ukrainian (UA) or English (EN).
For each model, the first row shows the best manual prompt result (from Table~\ref{tab:rq2-strategies}); the second row
shows the best optimized result. Bold marks the best result per column among API-accessed models.
$^\ddagger$\citet{gomez-etal-2023-low}.}
\label{tab:rq3-apo}
\end{table*}
%\footnotetext{For Claude Sonnet~4.6, minimal-edits + zero-shot outperforms minimal-edits + few-shot by 0.18 $F_{0.5}$ points; see Table~\ref{tab:rq2-strategies}.}

\paragraph{RQ3: LLM-assisted prompt optimization (Table~\ref{tab:rq3-apo}).}

\paragraph{Best results and gap to SOTA.}
The best optimized result is Gemini~3.1-Pro with minimal-edits + few-shot + optimized-v2 (\ref{prompt:apo};
$F_{0.5}=69.22$), followed closely by Gemini~3-Flash with the same prompt (68.43). This narrows the gap to fine-tuned
SOTA from 9.5 to 3.9~points. The gain on the target model is precision-driven: precision rises from 63.76 to 70.77
($+$7.0) while recall remains nearly unchanged (63.35 $\to$ 63.63). The remaining 3.9-point gap is concentrated in
precision (70.77 vs.\ 76.81), suggesting that closing it likely requires more focused instructions.

\paragraph{Improvement within Gemini.}
Since the minimal-edits + few-shot + optimized-v2 prompt (\ref{prompt:apo}) was tuned directly on Gemini~3-Flash, its
strong gain on that model ($F_{0.5}$: 61.38 $\to$ 68.43, $+$7.05) is expected. More notably, the same prompt transfers
successfully to Gemini~3.1-Pro, which achieves the overall best result ($F_{0.5}=69.22$), indicating that optimization
on a smaller model within the family can benefit larger variants.

\paragraph{Improvement on GPT.}
GPT models show mixed results. GPT-4.1-mini gains $+$6.87 points (47.97 $\to$ 54.84), a substantial improvement but
still the weakest absolute result. GPT-5.4 gains only $+$1.38 (59.69 $\to$ 61.07), suggesting that the stronger model
already captures most correction patterns encoded in the optimized prompt.

\paragraph{No improvement on Claude.}
The optimized prompt degrades both Claude models: Opus drops by $-$5.75 points (63.73 $\to$ 57.98) and Sonnet by $-$1.77
points (59.06 $\to$ 57.29), driven largely by a recall collapse (Opus: 49.75 $\to$ 38.27; Sonnet: 48.55 $\to$ 36.58).
Claude appears to interpret the optimized rules too conservatively, suppressing genuine corrections alongside false
positives. This finding shows that optimization on one model family does not necessarily guarantee improvement on
another.

% ===== FIGURE 1: RQ4 — Error Types =====
\begin{figure}[t!]
  \centering
  \includegraphics[width=\columnwidth]{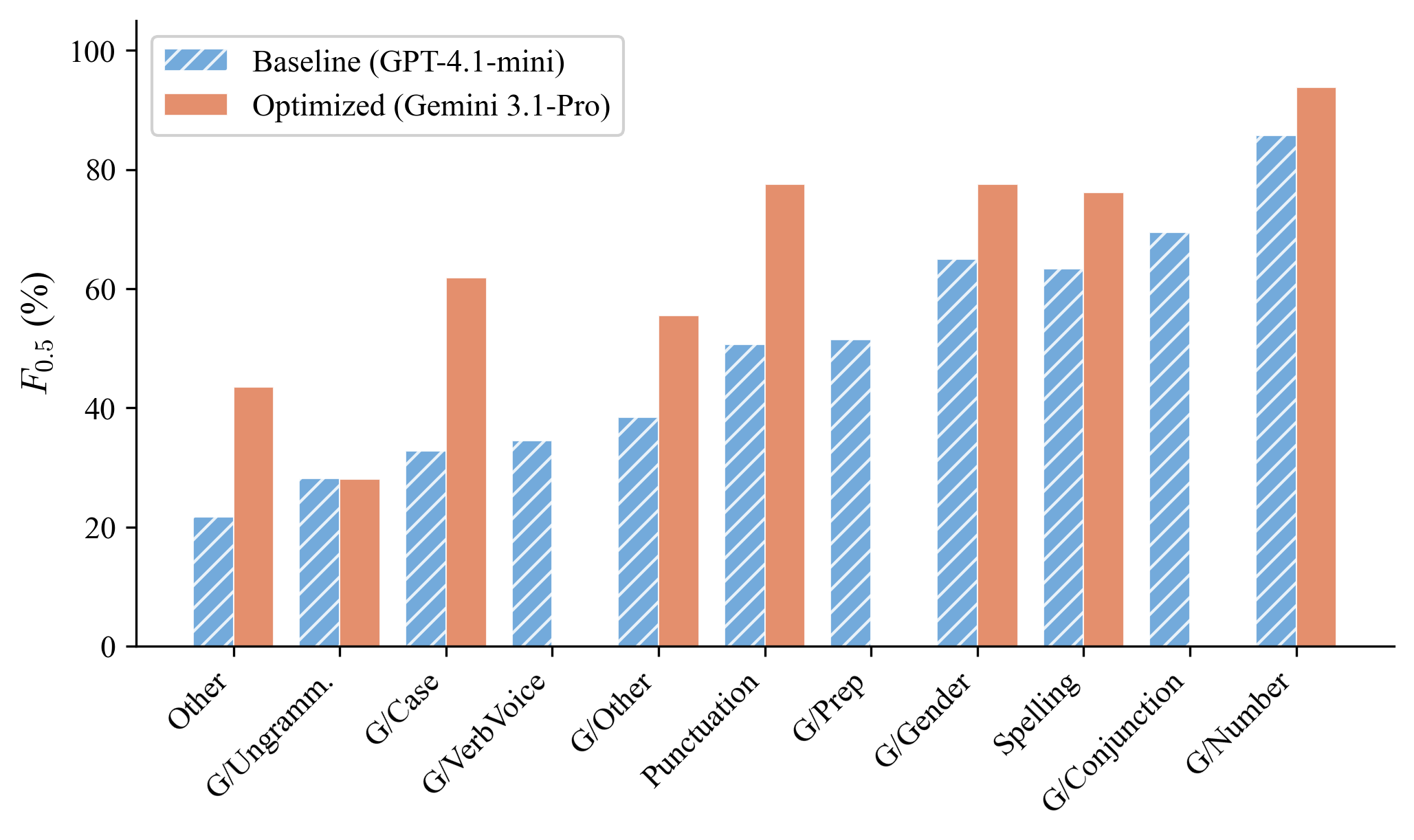}
  \caption{RQ4: Per-error-type $F_{0.5}$ on the UNLP 2023 test set. Baseline: GPT-4.1-mini (zero-shot
  (\ref{prompt:zero-shot-en}), EN); Optimized: Gemini~3.1-Pro (minimal-edits + few-shot + optimized-v2
  (\ref{prompt:apo}), UA). Error types sorted as in Table~\ref{tab:rq4-overcorrection} (by baseline overcorrection
  ratio, descending).}
  \label{fig:rq4-errors}
\end{figure}

\begin{table}[t!]
  \centering
  \footnotesize
  \setlength{\tabcolsep}{3pt}
  \begin{tabular}{lrr@{}}
  \toprule
  \textbf{Error Type} & \textbf{GPT-4.1-mini} & \textbf{Gemini~3.1-Pro} \\
  \midrule
  Other            & 3.67 & 0.50 \\
  G/Ungramm.       & 2.50 & 2.00 \\
  G/Case           & 1.88 & 0.39 \\
  G/VerbVoice      & 1.50 & -- \\
  G/Other          & 1.00 & 0.00 \\
  Punctuation      & 0.95 & 0.26 \\
  G/Prep           & 0.86 & -- \\
  G/Gender         & 0.60 & 0.22 \\
  Spelling         & 0.59 & 0.25 \\
  G/Conjunction    & 0.40 & $\infty$ \\
  G/Number         & 0.17 & 0.00 \\
  \bottomrule
  \end{tabular}
  \caption{RQ4: Overcorrection ratio (FP/TP) per error type; lower is better. The two columns compare the
  weakest (GPT-4.1-mini, zero-shot (\ref{prompt:zero-shot-en}), EN) and strongest (Gemini~3.1-Pro, minimal-edits +
  few-shot + optimized-v2 (\ref{prompt:apo}), UA) configurations from
  Tables~\ref{tab:rq2-strategies}--\ref{tab:rq3-apo}; note that both model and prompt differ. Rows sorted by
  GPT-4.1-mini ratio (descending); overcorrection (more false positives than true positives) occurs above 1.0. ``--''
  indicates no predictions for that type; $\infty$ indicates only false positives.}
  \label{tab:rq4-overcorrection}
\end{table}

\paragraph{RQ4: Where do minimal-edits instructions help and where do they fail? (Figure~\ref{fig:rq4-errors},
Table~\ref{tab:rq4-overcorrection}).}
We compare per-error-type $F_{0.5}$ between a standard zero-shot prompt (GPT-4.1-mini) and our best optimized prompt
(Gemini~3.1-Pro, minimal-edits + few-shot + optimized-v2; \ref{prompt:apo}) to identify which error categories benefit
most from detailed minimal-edits instructions. Note that this comparison reflects the combined effect of model choice,
prompt strategy, and prompt language; we select these two configurations as the weakest and strongest endpoints of our
evaluation pipeline.

\paragraph{Where minimal-edit instructions help.}
The largest $F_{0.5}$ gains appear in categories amenable to explicit rules. Punctuation improves from 50.67 to 77.56
($+$26.9), G/Case from 32.82 to 61.83 ($+$29.0), and G/Gender from 64.94 to 77.59 ($+$12.7). The overcorrection ratios
in Table~\ref{tab:rq4-overcorrection} confirm the mechanism: FP/TP drops from 0.95 to 0.26 for Punctuation, from 1.88 to
0.39 for G/Case, and from 0.60 to 0.22 for G/Gender. Spelling and G/Number are reliable under both configurations.

\paragraph{Where minimal-edit instructions fail.}
Three categories drop to $F_{0.5}=0$ under the optimized prompt: G/Prep, G/VerbVoice, and G/Conjunction. The baseline
achieves non-trivial $F_{0.5}$ scores on these types (51.47, 34.48, and 69.44), but the detailed minimal-edits rules
cause the model to avoid these corrections entirely. G/UngrammaticalStructure remains persistently overcorrected in both
settings ($F_{0.5}$: 28.17 $\to$ 28.04), indicating a structural difficulty that instructions cannot resolve.

\paragraph{Ukrainian-specific overcorrection patterns.}
Error analysis reveals five recurring patterns specific to Ukrainian, driven by the interaction between
English-calibrated correction heuristics and Ukrainian linguistic norms. Below, we report false positive counts out of
1{,}274 test sentences.

\textbf{En-dash over-normalization} (\textit{–}~$\to$~\textit{—};
{\raise.17ex\hbox{$\scriptstyle\sim$}}49~FP, 3.8\% of sentences).
Em-dashes are obligatory in direct speech but not elsewhere; the
model generalizes the rule indiscriminately.

\textbf{Dialogue reformatting}
({\raise.17ex\hbox{$\scriptstyle\sim$}}40~FP, 3.1\% of sentences).
The model converts acceptable quote-style dialogue
(\foreignlanguage{ukrainian}{«текст», — сказав} `\,"text," --- said') to dash-style,
applying a real Ukrainian norm where none was required.
A single prohibition rule was sufficient to suppress this pattern.

\textbf{Synonym and register substitution}
({\raise.17ex\hbox{$\scriptstyle\sim$}}30~FP, 2.4\% of sentences).
Acceptable words are replaced with literary alternatives
(\foreignlanguage{ukrainian}{\textit{знаходиться}} `is located'~$\to$~\foreignlanguage{ukrainian}{\textit{перебуває}}
`is situated'), violating the
minimal-edit constraint.

\textbf{Euphonic preposition alternation} (\foreignlanguage{ukrainian}{\textit{в/у}} `v/u',
\foreignlanguage{ukrainian}{\textit{з/із/зі}} `z/iz/zi';
{\raise.17ex\hbox{$\scriptstyle\sim$}}17~FP, 1.3\% of sentences).
Ukrainian preposition choice is phonetically conditioned; the model both over- and under-corrects within the same
category.

\textbf{Collapse of morphological variants.}
Ukrainian admits multiple grammatically correct surface forms
(\foreignlanguage{ukrainian}{\textit{навчались}}/\foreignlanguage{ukrainian}{\textit{навчалися}} `studied',
\foreignlanguage{ukrainian}{\textit{їх}}/\foreignlanguage{ukrainian}{\textit{їхній}} `their'); the model collapses these
to a single
preferred form. This space of acceptable alternations is open-ended
and cannot be exhaustively covered by prompt examples.

These patterns share a common cause: the model's correction heuristics are calibrated to English, where most of these
alternations do not exist. Overall, our strongest prompt (Gemini~3.1-Pro, minimal-edits + few-shot + optimized-v2
(\ref{prompt:apo}), UA) significantly reduces overcorrection for high-frequency, rule-based categories
(Table~\ref{tab:rq4-overcorrection}), but at the cost of the model becoming too conservative on low-frequency
grammatical categories. 

\section{Conclusion}

We presented the first systematic evaluation of prompting strategies for minimal-edit Ukrainian GEC using API-accessed
LLMs. While fine-tuned models currently dominate GEC benchmarks, we show that prompting alone can be competitive. On the
UNLP 2023 benchmark, our best configuration (Gemini~3.1-Pro with LLM-assisted optimization) reaches $F_{0.5}=69.22$,
closing over 90\% of the gap between the previous API-accessed result of \citet{katinskaia-yangarber-2024-gpt} (GPT-3.5,
$F_{0.5}=27.4$) and the fine-tuned SOTA of \citet{gomez-etal-2023-low} (mT5-large, $F_{0.5}=73.14$).

Our findings yield four takeaways. First, for zero-shot and few-shot prompts, English is sufficient for most models;
only Claude benefits from Ukrainian prompts (RQ1). Our best overall results, however, use Ukrainian minimal-edits
prompts, as the language-specific rules they encode require Ukrainian to express precisely. Second, the minimal-edits strategy provides
the largest gains, outperforming both zero-shot and few-shot baselines across all models (RQ2). Third, LLM-assisted
prompt optimization yields further improvements on the model family it was optimized for, but does not transfer reliably
across families (RQ3). Fourth, minimal-edits instructions yield the largest per-category gains for punctuation and case
errors, but cause the model to abandon several low-frequency grammatical categories entirely, revealing a
precision-recall tradeoff inherent to detailed prompting (RQ4).

%As commercial and open-source LLMs continue to advance, we believe that prompting-only methods are well positioned to close the remaining gap to fine-tuned models for Ukrainian GEC.

\section*{Limitations}

Our study has several limitations:

\begin{enumerate}
\item We evaluate on a single benchmark (UNLP 2023 GEC-only); results may not generalize to other Ukrainian GEC datasets
or domains.
\item Although we include a single open-source model (Lapa~v0.1.2) as a reference point, we do not systematically
compare against open-weight models (e.g., Llama~3, Mistral, Lapa, MamayLM) that could be prompted or fine-tuned without API
costs, leaving this as future work.
\item API-accessed models are opaque and subject to unannounced updates, making exact reproducibility difficult.
\item Although the UNLP 2023 test gold annotations are held out, the upstream UA-GEC train and valid splits are
publicly available. Since UA-GEC train is the source of our few-shot exemplars, it is plausible that commercial
LLMs encountered similar sentences and annotation patterns during pretraining, which could inflate recall on a
corpus sharing the same annotation conventions. A cleaner control would draw exemplars from an independent Ukrainian
GEC corpus, but no comparable dataset currently exists, so we treat our results as establishing initial prompting
baselines; prior API-accessed results for Ukrainian GEC at this scale are essentially absent.
\item We run each configuration only once; since LLM outputs are not fully deterministic, reproduced scores may differ/cos
slightly.
\item Our LLM-assisted prompt optimization pipeline optimizes $F_{0.5}$ on the development set, which may overfit to its
error distribution.
\item All our experiments are evaluated with span-based $F_{0.5}$ computed by ERRANT, the official metric of the UNLP
2023 shared task \cite{syvokon-romanyshyn-2023-unlp}; this differs from GLEU used in MultiGEC-2025
\cite{masciolini-etal-2025-multigec}, so scores are not directly comparable across benchmarks.
\item Optimized prompts are substantially longer (roughly 32$\times$ the zero-shot baseline; see
Figure~\ref{fig:prompt-tokens}), which may increase token cost and latency. Prompt caching, now widely supported by
providers, amortizes much of this overhead, making the net cost hard to estimate.
\end{enumerate}

\section*{Ethical Considerations}

In accordance with the conference policy on AI-based writing assistance, we disclose that ChatGPT, Claude, Gemini, and Grammarly were used for drafting, editing, and proofreading. All AI-generated text was reviewed by the authors,
who take full responsibility for the final content.

\section*{Acknowledgments}

We are deeply grateful to YouScan for fostering an inspiring environment that encourages both research and professional development. We also express our appreciation to the Faculty of Applied Sciences at the Ukrainian Catholic University for supporting this work as part of an M.Sc. thesis program. We gratefully acknowledge Mariana Romanyshyn and Oleksiy Syvokon for their assistance with the UNLP 2023 Shared Task. Finally, we extend our sincere gratitude to the anonymous reviewers for their insightful feedback and dedicated efforts in refining this manuscript.

\bibliography{custom}

\clearpage
\appendix
\nolinenumbers

\section{Prompts}
\label{sec:appendix-prompts}

Below we list all prompts in both English (\textsc{en}) and Ukrainian (\textsc{ua}) versions. The placeholder
\texttt{\{sentence\}} is replaced with the input sentence at inference time.

% ============================================================
\subsection{Zero-shot prompts}
\label{prompt:zero-shot}

The baseline prompt provides only a task description with no examples.

\subsubsection{Zero-shot prompt (EN)}
\label{prompt:zero-shot-en}
\begin{quote}\small
\texttt{Reply with a corrected version of the sentence with all grammatical and spelling errors fixed.}\\
\texttt{If there are no errors, reply with a copy of the original sentence.}\\
\texttt{Input sentence: <input\_text>}\\
\texttt{Corrected sentence:}\\
\end{quote}

\subsubsection{Zero-shot prompt (UA)}
\label{prompt:zero-shot-ua}
\begin{quote}\small
\foreignlanguage{ukrainian}{\texttt{Надай виправлену версію речення з виправленими всіма граматичними та орфографічними
помилками.}}\\
\foreignlanguage{ukrainian}{\texttt{Якщо помилок немає, надай копію оригінального речення.}}\\
\foreignlanguage{ukrainian}{\texttt{Вхідне речення: <input\_text>}}\\
\foreignlanguage{ukrainian}{\texttt{Виправлене речення:}}\\[4pt]
\textit{(English: `Provide a corrected version of the sentence with all grammatical and spelling errors fixed. If there
are no errors, provide a copy of the original sentence. Input sentence: <input\_text>. Corrected sentence:')}
\end{quote}

% ============================================================
\subsection{Few-shot prompts}
\label{prompt:few-shot}

The few-shot prompt prepends source--target pairs selected from the training set to cover spelling, punctuation, and
morphological error types.

\subsubsection{Few-shot prompt (EN)}
\label{prompt:few-shot-en}
  \begin{quote}\small
    \texttt{[Same header as Prompt~1]}

  \texttt{Examples:}

  \texttt{Input:} \foreignlanguage{ukrainian}{\texttt{Так само потерпає Україна і сьогодні від того що насправді
  талановитим людям заважають працювати усілякі посередності "у руля".}}\\
  \texttt{Output:} \foreignlanguage{ukrainian}{\texttt{Так само потерпає Україна і сьогодні від того, що насправді
  талановитим людям заважають працювати усілякі посередності "у руля".}}

  \texttt{Input:} \foreignlanguage{ukrainian}{\texttt{Це пов'язано з тим, що такі колективні рухи молекул води сильно
  збільшують характерні часи процесів які відбуваються в системі.}}\\
  \texttt{Output:} \foreignlanguage{ukrainian}{\texttt{Це пов'язано з тим, що такі колективні рухи молекул води сильно
  збільшують характерні часи процесів, які відбуваються в системі.}}

  \texttt{Input:} \foreignlanguage{ukrainian}{\texttt{Я ніколи не навчався у медичному коледжі, кажу, - Я лиш підійшов
  як звичайна людина подивитись чи можу бути чимось корисний.}}\\
  \texttt{Output:} \foreignlanguage{ukrainian}{\texttt{Я ніколи не навчався у медичному коледжі, кажу, - Я лиш підійшов
  як звичайна людина подивитись, чи можу бути чимось корисний.}}

  \texttt{Input:} \foreignlanguage{ukrainian}{\texttt{Це у місті швидка приїжджає, забирає хворого і везе у лікарню;
  якщо ж до лікарні кілька годин льоту то хворого можна і не довезти.}}\\
  \texttt{Output:} \foreignlanguage{ukrainian}{\texttt{Це у місті швидка приїжджає, забирає хворого і везе у лікарню;
  якщо ж до лікарні кілька годин льоту, то хворого можна і не довезти.}}

  \texttt{Input:} \foreignlanguage{ukrainian}{\texttt{Найбільше він любив тримати в руках старанно орнаментовані
  стародавні шолом і меч, котрі своїм золотом, здається гріли старого.}}\\
  \texttt{Output:} \foreignlanguage{ukrainian}{\texttt{Найбільше він любив тримати в руках старанно орнаментовані
  стародавні шолом і меч, котрі своїм золотом, здається, гріли старого.}}

  \texttt{Input:} \foreignlanguage{ukrainian}{\texttt{- Я часто казав тобі, що ти дурненька, - сказав він.}}\\
  \texttt{Output:} \foreignlanguage{ukrainian}{\texttt{— Я часто казав тобі, що ти дурненька, — сказав він.}}

  \texttt{Input:} \foreignlanguage{ukrainian}{\texttt{Така традиція також походить з Візантії, прикладом є зображення
  Андроніка II Палеолога.}}\\
  \texttt{Output:} \foreignlanguage{ukrainian}{\texttt{Така традиція також походить із Візантії, прикладом є зображення
  Андроніка II Палеолога.}}

  \texttt{Input:} \foreignlanguage{ukrainian}{\texttt{Як і більшість ділових людей, він не знав напамять жодного вірша і
  не памятав жодної казки, а тому щоразу мусив імпровізувати.}}\\
  \texttt{Output:} \foreignlanguage{ukrainian}{\texttt{Як і більшість ділових людей, він не знав напам'ять жодного вірша
  і не памятав жодної казки, а тому щоразу мусив імпровізувати.}}

  \texttt{Input:} \foreignlanguage{ukrainian}{\texttt{Настя, привіт! я хотіла уточнити про завдання Олі.}}\\
  \texttt{Output:} \foreignlanguage{ukrainian}{\texttt{Насте, привіт! Я хотіла уточнити про завдання Олі.}}

  \texttt{Input:} \foreignlanguage{ukrainian}{\texttt{Я принесу твої улюблені солодощі та обніму тебе міцно-міцно.}}\\
  \texttt{Output:} \foreignlanguage{ukrainian}{\texttt{Я принесу твої улюблені солодощі та обніму тебе міцно-міцно.}}

  \texttt{Input:} \foreignlanguage{ukrainian}{\texttt{Смакота ще та, скажу я вам))}}\\
  \texttt{Output:} \foreignlanguage{ukrainian}{\texttt{Смакота ще та, скажу я вам))}}

  \texttt{Input:} \foreignlanguage{ukrainian}{\texttt{У той час глибокий сенс народної мудрості нам був ще недоступним
  через брак досвіду.}}\\
  \texttt{Output:} \foreignlanguage{ukrainian}{\texttt{У той час глибокий сенс народної мудрості нам був ще недоступним
  через брак досвіду.}}

  \texttt{Input:} \foreignlanguage{ukrainian}{\texttt{Каналізація в будинках еволюціонує повільно, але все ж таки
  змінюється.}}\\
  \texttt{Output:} \foreignlanguage{ukrainian}{\texttt{Каналізація в будинках еволюціонує повільно, але все ж таки
  змінюється.}}

  \texttt{Input sentence: \{input\_text\}}\\
\texttt{Corrected sentence:}\\
    \end{quote}

\subsubsection{Few-shot prompt (UA)}
\label{prompt:few-shot-ua}
\begin{quote}\small
\foreignlanguage{ukrainian}{\texttt{[Той самий заголовок, що і в Prompt~1]}} \textit{(English: `[Same header as
Prompt~1]')}

  \foreignlanguage{ukrainian}{\texttt{Приклади:}} \textit{(`Examples:')}

\foreignlanguage{ukrainian}{\texttt{Вхід: Так само потерпає Україна і сьогодні від того що насправді талановитим людям
заважають працювати усілякі посередності "у руля".}}\\
  \foreignlanguage{ukrainian}{\texttt{Вихід: Так само потерпає Україна і сьогодні від того, що насправді талановитим
  людям заважають працювати усілякі посередності "у руля".}}\\[1mm]

  \foreignlanguage{ukrainian}{\texttt{Вхід: Я ніколи не навчався у медичному коледжі, кажу, - Я лиш підійшов як звичайна
  людина подивитись чи можу бути чимось корисний.}}\\
  \foreignlanguage{ukrainian}{\texttt{Вихід: Я ніколи не навчався у медичному коледжі, кажу, - Я лиш підійшов як
  звичайна людина подивитись, чи можу бути чимось корисний.}}\\[1mm]

  \foreignlanguage{ukrainian}{\texttt{Вхід: Це у місті швидка приїжджає, забирає хворого і везе у лікарню; якщо ж до
  лікарні кілька годин льоту то хворого можна і не довезти.}}\\
  \foreignlanguage{ukrainian}{\texttt{Вихід: Це у місті швидка приїжджає, забирає хворого і везе у лікарню; якщо ж до
  лікарні кілька годин льоту, то хворого можна і не довезти.}}\\[1mm]

  \foreignlanguage{ukrainian}{\texttt{Вхід: Найбільше він любив тримати в руках старанно орнаментовані стародавні шолом
  і меч, котрі своїм золотом, здається гріли старого.}}\\
  \foreignlanguage{ukrainian}{\texttt{Вихід: Найбільше він любив тримати в руках старанно орнаментовані стародавні шолом
  і меч, котрі своїм золотом, здається, гріли старого.}}\\[1mm]

  \foreignlanguage{ukrainian}{\texttt{Вхід: - Я часто казав тобі, що ти дурненька, - сказав він.}}\\
  \foreignlanguage{ukrainian}{\texttt{Вихід: — Я часто казав тобі, що ти дурненька, — сказав він.}}\\[1mm]

  \foreignlanguage{ukrainian}{\texttt{Вхід: Така традиція також походить з Візантії, прикладом є зображення Андроніка II
  Палеолога.}}\\
  \foreignlanguage{ukrainian}{\texttt{Вихід: Така традиція також походить із Візантії, прикладом є зображення Андроніка
  II Палеолога.}}\\[1mm]

  \foreignlanguage{ukrainian}{\texttt{Вхід: Як і більшість ділових людей, він не знав напамять жодного вірша і не
  памятав жодної казки, а тому щоразу мусив імпровізувати.}}\\
  \foreignlanguage{ukrainian}{\texttt{Вихід: Як і більшість ділових людей, він не знав напам'ять жодного вірша і не
  памятав жодної казки, а тому щоразу мусив імпровізувати.}}\\[1mm]

  \foreignlanguage{ukrainian}{\texttt{Вхід: Настя, привіт! я хотіла уточнити про завдання Олі.}}\\
  \foreignlanguage{ukrainian}{\texttt{Вихід: Насте, привіт! Я хотіла уточнити про завдання Олі.}}\\[1mm]

  \foreignlanguage{ukrainian}{\texttt{Вхід: Я принесу твої улюблені солодощі та обніму тебе міцно-міцно.}}\\
  \foreignlanguage{ukrainian}{\texttt{Вихід: Я принесу твої улюблені солодощі та обніму тебе міцно-міцно.}}\\[1mm]

  \foreignlanguage{ukrainian}{\texttt{Вхід: Смакота ще та, скажу я вам))}}\\
  \foreignlanguage{ukrainian}{\texttt{Вихід: Смакота ще та, скажу я вам))}}\\[1mm]

  \foreignlanguage{ukrainian}{\texttt{Вхід: У той час глибокий сенс народної мудрості нам був ще недоступним через брак
  досвіду.}}\\
  \foreignlanguage{ukrainian}{\texttt{Вихід: У той час глибокий сенс народної мудрості нам був ще недоступним через брак
  досвіду.}}\\[1mm]

  \foreignlanguage{ukrainian}{\texttt{Вхід: Каналізація в будинках еволюціонує повільно, але все ж таки змінюється.}}\\
  \foreignlanguage{ukrainian}{\texttt{Вихід: Каналізація в будинках еволюціонує повільно, але все ж таки
  змінюється.}}\\[2mm]

  \foreignlanguage{ukrainian}{\texttt{Вхідне речення: \{input\_text\}}}\\
  \foreignlanguage{ukrainian}{\texttt{Виправлене речення:}}\\[4pt]
\textit{(English:
\foreignlanguage{ukrainian}{\texttt{Вхід}}/\foreignlanguage{ukrainian}{\texttt{Вихід}}~=~`Input'/`Output';
\foreignlanguage{ukrainian}{\texttt{Вхідне речення}}~=~`Input sentence'; \foreignlanguage{ukrainian}{\texttt{Виправлене
речення}}~=~`Corrected sentence'. The few-shot examples are the same Ukrainian GEC sentence pairs as in the EN variant
(\ref{prompt:few-shot-en}), with Ukrainian keywords.)}

  \end{quote}

% ============================================================
\subsection{Minimal-edits zero-shot prompts}
\label{prompt:minimal-edits-zero-shot}

This prompt replaces the generic system prompt with a detailed minimal-edit instruction containing Ukrainian-specific
grammar rules. No few-shot examples are included.

\subsubsection{Minimal-edits zero-shot prompt (UA)}
\label{prompt:minimal-edits-zero-shot-ua}

\begin{quote}\small
  \foreignlanguage{ukrainian}{\texttt{Ти — система виправлення українських граматичних помилок. Внось МІНІМАЛЬНІ зміни,
  щоб виправити ЛИШЕ явні граматичні, орфографічні та пунктуаційні помилки. НЕ
  переписуй, не перефразовуй і не замінюй слова синонімами. Точно зберігай оригінальне формулювання.}}\\[1mm]

  \foreignlanguage{ukrainian}{\texttt{Виправляй ЛИШЕ такі типи помилок:}}\\[1mm]
  \foreignlanguage{ukrainian}{\texttt{1. Орфографія: явні орфографічні помилки (друкарські помилки, неправильні
  літери).}}\\
  \foreignlanguage{ukrainian}{\texttt{2. Пунктуація: пропущені або зайві коми, крапки, знаки питання; використання тире
  (—) замість дефіса (-) у діалогах та вставних конструкціях.}}\\
  \foreignlanguage{ukrainian}{\texttt{3. G/Case: некоректне вживання відмінкової форми (зокрема кличний відмінок при
  звертаннях).}}\\
  \foreignlanguage{ukrainian}{\texttt{4. G/Gender: некоректне вживання форми роду.}}\\
  \foreignlanguage{ukrainian}{\texttt{5. G/Number: некоректне вживання форми числа.}}\\
  \foreignlanguage{ukrainian}{\texttt{6. G/Aspect: некоректне вживання форми виду дієслова.}}\\
  \foreignlanguage{ukrainian}{\texttt{7. G/Tense: некоректне вживання часової форми дієслова.}}\\
  \foreignlanguage{ukrainian}{\texttt{8. G/VerbVoice: некоректне вживання форми стану дієслова.}}\\
  \foreignlanguage{ukrainian}{\texttt{9. G/PartVoice: некоректне вживання форми стану дієприкметника.}}\\
  \foreignlanguage{ukrainian}{\texttt{10. G/VerbAForm: некоректне вживання аналітичної форми дієслова.}}\\
  \foreignlanguage{ukrainian}{\texttt{11. G/Prep: некоректне вживання прийменника.}}\\
  \foreignlanguage{ukrainian}{\texttt{12. G/Participle: некоректне вживання дієприслівника.}}\\
  \foreignlanguage{ukrainian}{\texttt{13. G/UngrammaticalStructure: порушення граматичних норм у синтаксичних
  конструкціях.}}\\
  \foreignlanguage{ukrainian}{\texttt{14. G/Comparison: некоректна форма ступенів порівняння.}}\\
  \foreignlanguage{ukrainian}{\texttt{15. G/Conjunction: некоректне вживання сполучників.}}\\
  \foreignlanguage{ukrainian}{\texttt{16. G/Other: інші граматичні помилки.}}\\[1mm]

  \foreignlanguage{ukrainian}{\texttt{ВАЖЛИВІ ПРАВИЛА УКРАЇНСЬКОЇ МОВИ:}}\\
  \foreignlanguage{ukrainian}{\texttt{- Прийменник «у» вживається перед приголосними (у школі, у місті, у готелі), «в» —
  перед голосними та на початку речення.}}\\
  \foreignlanguage{ukrainian}{\texttt{- Прийменник «об» вживається перед голосними (об одинадцятій), «о» — перед
  приголосними.}}\\
  \foreignlanguage{ukrainian}{\texttt{- У діалогах вживається тире (—), а не дефіс (-): «Текст», — сказав він. — Текст
  далі.}}\\
  \foreignlanguage{ukrainian}{\texttt{- Вставні слова (може, мабуть, звичайно, здається) виділяються комами з обох
  боків.}}\\
  \foreignlanguage{ukrainian}{\texttt{- Кличний відмінок при звертаннях: Настя → Насте, Олег → Олеже, мама →
  мамо.}}\\[1mm]

  \foreignlanguage{ukrainian}{\texttt{СУВОРІ ПРАВИЛА:}}\\
  \foreignlanguage{ukrainian}{\texttt{- Виправляй ЛИШЕ явні помилки, перелічені вище}}\\
  \foreignlanguage{ukrainian}{\texttt{- Для кожної помилки внось НАЙМЕНШУ можливу зміну}}\\
  \foreignlanguage{ukrainian}{\texttt{- НІКОЛИ не замінюй слова синонімами і не перефразовуй (залишай «буду йти», НЕ
  змінюй на «піду»)}}\\
  \foreignlanguage{ukrainian}{\texttt{- НІКОЛИ не змінюй слово на інше, якщо воно не аписане з помилкою}}\\
  \foreignlanguage{ukrainian}{\texttt{- Зберігай оригінальний стиль лапок (" або «»), НЕ перетворюй один тип лапок на
  інший}}\\
  \foreignlanguage{ukrainian}{\texttt{- Зберігай оригінальне використання великих/малих літер, якщо це не явна
  помилка}}\\
  \foreignlanguage{ukrainian}{\texttt{- Якщо граматична форма є прийнятною, залишай її, навіть якщо можлива й інша
  форма}}\\
  \foreignlanguage{ukrainian}{\texttt{- Якщо є сумнів, НЕ змінюй}}\\
  \foreignlanguage{ukrainian}{\texttt{- Якщо помилок немає, повертай оригінальний текст БЕЗ ЗМІН}}\\
  \foreignlanguage{ukrainian}{\texttt{- Поверни ЛИШЕ виправлений текст}}
  \end{quote}

\noindent\textit{English translation of the above prompt:}
\begin{quote}\small\itshape
You are a Ukrainian grammatical error correction system. Make MINIMAL changes to fix ONLY obvious grammatical, spelling,
and punctuation errors. Do NOT rewrite, rephrase, or substitute synonyms. Preserve the original wording exactly.\\[1mm]
Fix ONLY the following types of errors:\\
1. Spelling: obvious spelling errors (typos, wrong letters).\\
2. Punctuation: missing or extra commas, periods, question marks; use of em-dash (---) instead of hyphen (-) in
dialogues and parenthetical constructions.\\
3. G/Case: incorrect case form (especially vocative case in forms of address).\\
4. G/Gender: incorrect gender form.\\
5. G/Number: incorrect number form.\\
6. G/Aspect: incorrect verb aspect form.\\
7. G/Tense: incorrect verb tense form.\\
8. G/VerbVoice: incorrect verb voice form.\\
9. G/PartVoice: incorrect participle voice form.\\
10. G/VerbAForm: incorrect analytical verb form.\\
11. G/Prep: incorrect preposition usage.\\
12. G/Participle: incorrect adverbial participle usage.\\
13. G/UngrammaticalStructure: grammatical norm violations in syntactic constructions.\\
14. G/Comparison: incorrect comparative/superlative form.\\
15. G/Conjunction: incorrect conjunction usage.\\
16. G/Other: other grammatical errors.\\[1mm]
IMPORTANT RULES OF UKRAINIAN:\\
- Preposition ``u'' is used before consonants (u shkoli, u misti), ``v'' before vowels and at sentence start.\\
- Preposition ``ob'' is used before vowels (ob odynnadtsiatii), ``o'' before consonants.\\
- Em-dash (---) is used in dialogues, not hyphen (-).\\
- Parenthetical words (maybe, probably, of course, it seems) are set off by commas on both sides.\\
- Vocative case in forms of address: Nastia $\to$ Naste, Oleh $\to$ Olezhe, mama $\to$ mamo.\\[1mm]
STRICT RULES:\\
- Fix ONLY obvious errors listed above\\
- For each error, make the SMALLEST possible change\\
- NEVER substitute synonyms or rephrase\\
- NEVER change a word to another unless it is misspelled\\
- Preserve original quote style, do NOT convert one type to another\\
- Preserve original capitalization unless it is a clear error\\
- If a grammatical form is acceptable, leave it even if another form is possible\\
- If in doubt, do NOT change\\
- If there are no errors, return the original text WITHOUT CHANGES\\
- Return ONLY the corrected text
\end{quote}

% ============================================================
\subsection{Minimal-edits few-shot prompts}
\label{prompt:minimal-edits-few-shot}

This prompt combines the minimal-edits system prompt (Appendix~\ref{prompt:minimal-edits-zero-shot}) with few-shot
examples from the training set.

\subsubsection{Minimal-edits few-shot prompt (UA)}
\label{prompt:minimal-edits-few-shot-ua}
\begin{quote}\small
  \foreignlanguage{ukrainian}{\texttt{Ти — система виправлення українських граматичних помилок. Внось МІНІМАЛЬНІ зміни,
  щоб виправити ЛИШЕ явні граматичні, орфографічні та пунктуаційні помилки. НЕ
  переписуй, не перефразовуй і не замінюй слова синонімами. Точно зберігай оригінальне формулювання.}}\\[1mm]

  \foreignlanguage{ukrainian}{\texttt{Виправляй ЛИШЕ такі типи помилок:}}\\[1mm]
  \foreignlanguage{ukrainian}{\texttt{1. Орфографія: явні орфографічні помилки (друкарські помилки, неправильні
  літери).}}\\
  \foreignlanguage{ukrainian}{\texttt{2. Пунктуація: пропущені або зайві коми, крапки, знаки питання; використання тире
  (—) замість дефіса (-) у діалогах та вставних конструкціях.}}\\
  \foreignlanguage{ukrainian}{\texttt{3. G/Case: некоректне вживання відмінкової форми (зокрема кличний відмінок при
  звертаннях).}}\\
  \foreignlanguage{ukrainian}{\texttt{4. G/Gender: некоректне вживання форми роду.}}\\
  \foreignlanguage{ukrainian}{\texttt{5. G/Number: некоректне вживання форми числа.}}\\
  \foreignlanguage{ukrainian}{\texttt{6. G/Aspect: некоректне вживання форми виду дієслова.}}\\
  \foreignlanguage{ukrainian}{\texttt{7. G/Tense: некоректне вживання часової форми дієслова.}}\\
  \foreignlanguage{ukrainian}{\texttt{8. G/VerbVoice: некоректне вживання форми стану дієслова.}}\\
  \foreignlanguage{ukrainian}{\texttt{9. G/PartVoice: некоректне вживання форми стану дієприкметника.}}\\
  \foreignlanguage{ukrainian}{\texttt{10. G/VerbAForm: некоректне вживання аналітичної форми дієслова.}}\\
  \foreignlanguage{ukrainian}{\texttt{11. G/Prep: некоректне вживання прийменника.}}\\
  \foreignlanguage{ukrainian}{\texttt{12. G/Participle: некоректне вживання дієприслівника.}}\\
  \foreignlanguage{ukrainian}{\texttt{13. G/UngrammaticalStructure: порушення граматичних норм у синтаксичних
  конструкціях.}}\\
  \foreignlanguage{ukrainian}{\texttt{14. G/Comparison: некоректна форма ступенів порівняння.}}\\
  \foreignlanguage{ukrainian}{\texttt{15. G/Conjunction: некоректне вживання сполучників.}}\\
  \foreignlanguage{ukrainian}{\texttt{16. G/Other: інші граматичні помилки.}}\\[1mm]

  \foreignlanguage{ukrainian}{\texttt{ВАЖЛИВІ ПРАВИЛА УКРАЇНСЬКОЇ МОВИ:}}\\
  \foreignlanguage{ukrainian}{\texttt{- Прийменник «у» вживається перед приголосними (у школі, у місті, у готелі), «в» —
  перед голосними та на початку речення.}}\\
  \foreignlanguage{ukrainian}{\texttt{- Прийменник «об» вживається перед голосними (об одинадцятій), «о» — перед
  приголосними.}}\\
  \foreignlanguage{ukrainian}{\texttt{- У діалогах вживається тире (—), а не дефіс (-): «Текст», — сказав він. — Текст
  далі.}}\\
  \foreignlanguage{ukrainian}{\texttt{- Вставні слова (може, мабуть, звичайно, здається) виділяються комами з обох
  боків.}}\\
  \foreignlanguage{ukrainian}{\texttt{- Кличний відмінок при звертаннях: Настя → Насте, Олег → Олеже, мама →
  мамо.}}\\[1mm]

  \foreignlanguage{ukrainian}{\texttt{СУВОРІ ПРАВИЛА:}}\\
  \foreignlanguage{ukrainian}{\texttt{- Виправляй ЛИШЕ явні помилки, перелічені вище}}\\
  \foreignlanguage{ukrainian}{\texttt{- Для кожної помилки внось НАЙМЕНШУ можливу зміну}}\\
  \foreignlanguage{ukrainian}{\texttt{- НІКОЛИ не замінюй слова синонімами і не перефразовуй (залишай «буду йти», НЕ
  змінюй на «піду»)}}\\
  \foreignlanguage{ukrainian}{\texttt{- НІКОЛИ не змінюй слово на інше, якщо воно не аписане з помилкою}}\\
  \foreignlanguage{ukrainian}{\texttt{- Зберігай оригінальний стиль лапок (" або «»), НЕ перетворюй один тип лапок на
  інший}}\\
  \foreignlanguage{ukrainian}{\texttt{- Зберігай оригінальне використання великих/малих літер, якщо це не явна
  помилка}}\\
  \foreignlanguage{ukrainian}{\texttt{- Якщо граматична форма є прийнятною, залишай її, навіть якщо можлива й інша
  форма}}\\
  \foreignlanguage{ukrainian}{\texttt{- Якщо є сумнів, НЕ змінюй}}\\
  \foreignlanguage{ukrainian}{\texttt{- Якщо помилок немає, повертай оригінальний текст БЕЗ ЗМІН}}\\[1mm]

  \foreignlanguage{ukrainian}{\texttt{Приклади:}}\\[1mm]

  \foreignlanguage{ukrainian}{\texttt{Вхід: Так само потерпає Україна і сьогодні від того що насправді талановитим людям
  заважають працювати усілякі посередності "у руля".}}\\
  \foreignlanguage{ukrainian}{\texttt{Вихід: Так само потерпає Україна і сьогодні від того, що насправді талановитим
  людям заважають працювати усілякі посередності "у руля".}}\\[1mm]

  \foreignlanguage{ukrainian}{\texttt{Вхід: Я ніколи не навчався у медичному коледжі, кажу, - Я лиш підійшов як звичайна
  людина подивитись чи можу бути чимось корисний.}}\\
  \foreignlanguage{ukrainian}{\texttt{Вихід: Я ніколи не навчався у медичному коледжі, кажу, - Я лиш підійшов як
  звичайна людина подивитись, чи можу бути чимось корисний.}}\\[1mm]

  \foreignlanguage{ukrainian}{\texttt{Вхід: Це у місті швидка приїжджає, забирає хворого і везе у лікарню; якщо ж до
  лікарні кілька годин льоту то хворого можна і не довезти.}}\\
  \foreignlanguage{ukrainian}{\texttt{Вихід: Це у місті швидка приїжджає, забирає хворого і везе у лікарню; якщо ж до
  лікарні кілька годин льоту, то хворого можна і не довезти.}}\\[1mm]

  \foreignlanguage{ukrainian}{\texttt{Вхід: Найбільше він любив тримати в руках старанно орнаментовані стародавні шолом
  і меч, котрі своїм золотом, здається гріли старого.}}\\
  \foreignlanguage{ukrainian}{\texttt{Вихід: Найбільше він любив тримати в руках старанно орнаментовані стародавні шолом
  і меч, котрі своїм золотом, здається, гріли старого.}}\\[1mm]

  \foreignlanguage{ukrainian}{\texttt{Вхід: - Я часто казав тобі, що ти дурненька, - сказав він.}}\\
  \foreignlanguage{ukrainian}{\texttt{Вихід: — Я часто казав тобі, що ти дурненька, — сказав він.}}\\[1mm]

  \foreignlanguage{ukrainian}{\texttt{Вхід: Така традиція також походить з Візантії, прикладом є зображення Андроніка II
  Палеолога.}}\\
  \foreignlanguage{ukrainian}{\texttt{Вихід: Така традиція також походить із Візантії, прикладом є зображення Андроніка
  II Палеолога.}}\\[1mm]

  \foreignlanguage{ukrainian}{\texttt{Вхід: Як і більшість ділових людей, він не знав напамять жодного вірша і не
  памятав жодної казки, а тому щоразу мусив імпровізувати.}}\\
  \foreignlanguage{ukrainian}{\texttt{Вихід: Як і більшість ділових людей, він не знав напам'ять жодного вірша і не
  памятав жодної казки, а тому щоразу мусив імпровізувати.}}\\[1mm]

  \foreignlanguage{ukrainian}{\texttt{Вхід: Настя, привіт! я хотіла уточнити про завдання Олі.}}\\
  \foreignlanguage{ukrainian}{\texttt{Вихід: Насте, привіт! Я хотіла уточнити про завдання Олі.}}\\[1mm]

  \foreignlanguage{ukrainian}{\texttt{Вхід: Я принесу твої улюблені солодощі та обніму тебе міцно-міцно.}}\\
  \foreignlanguage{ukrainian}{\texttt{Вихід: Я принесу твої улюблені солодощі та обніму тебе міцно-міцно.}}\\[1mm]

  \foreignlanguage{ukrainian}{\texttt{Вхід: Смакота ще та, скажу я вам))}}\\
  \foreignlanguage{ukrainian}{\texttt{Вихід: Смакота ще та, скажу я вам))}}\\[1mm]

  \foreignlanguage{ukrainian}{\texttt{Вхід: У той час глибокий сенс народної мудрості нам був ще недоступним через брак
  досвіду.}}\\
  \foreignlanguage{ukrainian}{\texttt{Вихід: У той час глибокий сенс народної мудрості нам був ще недоступним через брак
  досвіду.}}\\[1mm]

  \foreignlanguage{ukrainian}{\texttt{Вхід: Каналізація в будинках еволюціонує повільно, але все ж таки змінюється.}}\\
  \foreignlanguage{ukrainian}{\texttt{Вихід: Каналізація в будинках еволюціонує повільно, але все ж таки
  змінюється.}}\\[2mm]

  \foreignlanguage{ukrainian}{\texttt{Вхідне речення: \{input\_text\}}}\\
  \foreignlanguage{ukrainian}{\texttt{Виправлене речення:}}
\end{quote}

\noindent\textit{English: The system prompt is identical to the minimal-edits zero-shot prompt
(\ref{prompt:minimal-edits-zero-shot-ua}); see the English translation there. The few-shot examples are the same
Ukrainian GEC sentence pairs as in the few-shot EN variant (\ref{prompt:few-shot-en}). Keywords:
\foreignlanguage{ukrainian}{\texttt{Приклади}}~=~`Examples';
\foreignlanguage{ukrainian}{\texttt{Вхід}}/\foreignlanguage{ukrainian}{\texttt{Вихід}}~=~`Input'/`Output'.}

% ============================================================
\subsection{Optimized prompts}
\label{sec:optimized-prompts}

The following prompts were produced by LLM-assisted prompt optimization (Section~\ref{sec:prompt-design}). Each was
derived from its parent prompt via iterative refinement on the development set (see Section~\ref{sec:prompt-design} for
the optimization procedure).

\subsubsection{Minimal-edits + few-shot + optimized-v1 (UA): optimized on GPT-4.1-mini}
\label{prompt:apo A002}

\begin{quote}\small
  \foreignlanguage{ukrainian}{\texttt{Ти — система виправлення українських граматичних помилок. Внось МІНІМАЛЬНІ зміни,
  щоб виправити ЛИШЕ безсумнівні граматичні, орфографічні та пунктуаційні помилки.}}\
  \
  \foreignlanguage{ukrainian}{\texttt{НЕ переписуй, не перефразовуй і не замінюй слова синонімами.}}\\[1mm]

  \foreignlanguage{ukrainian}{\texttt{Виправляй:}}\\
  \foreignlanguage{ukrainian}{\texttt{- Орфографічні помилки (друкарські помилки, пропущені/зайві літери, неправильне
  написання разом/окремо: незважати → не зважати, буд-якому → будь-якому).}}\\
  \foreignlanguage{ukrainian}{\texttt{- Пунктуацію: пропущені коми перед підрядними сполучниками (що, який, бо, чи,
  коли, щоб, де, поки), при звертаннях, при вставних словах (мабуть, може, звичайно). У
  прямій мові дефіс (-) замінюй на тире (—): "текст" - сказав → "текст", — сказав.}}\\
  \foreignlanguage{ukrainian}{\texttt{- Відмінкові помилки (зокрема кличний відмінок у звертаннях: Привіт Настя →
  Привіт, Насте).}}\\
  \foreignlanguage{ukrainian}{\texttt{- Узгодження роду, числа, відмінка в словосполученнях.}}\\
  \foreignlanguage{ukrainian}{\texttt{- Прийменники: о/об (об одинадцятій, о третій); з → зі/із перед збігом приголосних
  (з зображенням → зі зображенням, з Візантії → із Візантії).}}\\[1mm]

  \foreignlanguage{ukrainian}{\texttt{НЕ змінюй:}}\\
  \foreignlanguage{ukrainian}{\texttt{- НІКОЛИ не замінюй слова синонімами і не змінюй форми слів на альтернативні
  (виказував, кучею, достойною, вивести — залишай як є).}}\\
  \foreignlanguage{ukrainian}{\texttt{- НЕ переформатовуй діалоги: якщо діалог оформлений лапками ("/«»), зберігай
  лапки, НЕ замінюй їх на тире.}}\\
  \foreignlanguage{ukrainian}{\texttt{- НЕ змінюй граматичні форми, які є допустимими варіантами: відмінкові форми
  прикметників (недоступним/недоступний), варіанти дієслів (навчались/навчалися), форми
  займенників (їх/їхній) — якщо форма граматично допустима, залишай її.}}\\
  \foreignlanguage{ukrainian}{\texttt{- Стиль та тон тексту: неформальний текст (чати, смс) залишай як є — не додавай
  крапки в кінці, не прибирай смайлики )).}}\\
  \foreignlanguage{ukrainian}{\texttt{- Порядок слів у реченні.}}\\
  \foreignlanguage{ukrainian}{\texttt{- Великі/малі літери, крім початку речення після крапки.}}\\
  \foreignlanguage{ukrainian}{\texttt{- Лапки: зберігай оригінальний стиль.}}\\
  \foreignlanguage{ukrainian}{\texttt{- Розділові знаки кінця речення: НЕ змінюй . на ? або навпаки.}}\\
  \foreignlanguage{ukrainian}{\texttt{- НЕ додавай тире (—) там, де його не було в оригіналі, окрім прямої мови.}}\\
  \foreignlanguage{ukrainian}{\texttt{- Дефіс у складених словах та повторах (міцно-міцно, дере-дере-дере, все-таки, все
  ж таки — залишай як є).}}\\
  \foreignlanguage{ukrainian}{\texttt{- Якщо сумніваєшся — НЕ змінюй.}}

  \foreignlanguage{ukrainian}{\texttt{[Ті самі приклади, що і в Prompt~2]}} \textit{(`[Same examples as in Prompt~2]')}

  \foreignlanguage{ukrainian}{\texttt{Поверни ЛИШЕ виправлений текст.}} \textit{(`Return ONLY the corrected text.')}

  \end{quote}

\noindent\textit{English translation of the instruction part:}
\begin{quote}\small\itshape
You are a Ukrainian grammatical error correction system. Make MINIMAL changes to fix ONLY unambiguous grammatical,
spelling, and punctuation errors. Do NOT rewrite, rephrase, or substitute synonyms.\\[1mm]
Fix:\\
- Spelling errors (typos, missing/extra letters, incorrect joined/separate writing: nezvazhaty $\to$ ne zvazhaty,
bud-yakomu $\to$ bud\textquotesingle-yakomu).\\
- Punctuation: missing commas before subordinate conjunctions (shcho, yakyi, bo, chy, koly, shchob, de, poky), in forms
of address, with parenthetical words (mabut\textquotesingle, mozhe, zvychaino). In direct speech, replace hyphen (-)
with em-dash (---).\\
- Case errors (especially vocative in address: Pryvit Nastia $\to$ Pryvit, Naste).\\
- Gender, number, case agreement in phrases.\\
- Prepositions: o/ob; z $\to$ zi/iz before consonant clusters.\\[1mm]
Do NOT change:\\
- NEVER substitute synonyms or change word forms to alternatives --- leave as is.\\
- Do NOT reformat dialogues: if dialogue uses quotes, keep quotes, do NOT replace with dashes.\\
- Do NOT change grammatically acceptable variant forms.\\
- Text style and tone: leave informal text (chats, SMS) as is.\\
- Word order, capitalization (except after period), quote style, sentence-final punctuation.\\
- Do NOT add em-dashes where there were none, except in direct speech.\\
- Hyphens in compound words and repetitions --- leave as is.\\
- If in doubt --- do NOT change.
\end{quote}

\subsubsection{Minimal-edits + few-shot + optimized-v2 (UA): optimized on Gemini 3-Flash}
\label{prompt:apo}

The prompt below was derived from minimal-edits + few-shot + optimized-v1 (\ref{prompt:apo A002}) by further
LLM-assisted optimization on Gemini~3-Flash over 5 iterations on the development set. It includes additional
Ukrainian-specific rules discovered during optimization.

\begin{quote}\small
\foreignlanguage{ukrainian}{\texttt{Ти — система виправлення українських граматичних помилок. Внось МІНІМАЛЬНІ зміни,
щоб виправити ЛИШЕ безсумнівні граматичні, орфографічні та пунктуаційні помилки. НЕ
  переписуй, не перефразовуй і не замінюй слова синонімами.}}\\

  \foreignlanguage{ukrainian}{\texttt{Виправляй:}}\\
  \foreignlanguage{ukrainian}{\texttt{- Орфографічні помилки (друкарські помилки, пропущені/зайві літери, неправильне
  написання разом/окремо: незважати → не зважати, буд-якому → будь-якому).}}\\
  \foreignlanguage{ukrainian}{\texttt{- Пунктуацію: пропущені коми перед підрядними сполучниками (що, який, бо, чи,
  коли, щоб, де, поки), при звертаннях, при вставних словах (мабуть, може, звичайно). У
  прямій мові дефіс (-) замінюй на тире (—): "текст" - сказав → "текст", — сказав.}}\\
  \foreignlanguage{ukrainian}{\texttt{- Відмінкові помилки (зокрема кличний відмінок у звертаннях: Привіт Настя →
  Привіт, Насте).}}\\
  \foreignlanguage{ukrainian}{\texttt{- Узгодження роду, числа, відмінка в словосполученнях.}}\\
  \foreignlanguage{ukrainian}{\texttt{- Прийменники: о/об (об одинадцятій, о третій); з → зі/із перед збігом приголосних
  (з зображенням → зі зображенням, з Візантії → із Візантії).}}\\

  \foreignlanguage{ukrainian}{\texttt{НЕ змінюй:}}\\
  \foreignlanguage{ukrainian}{\texttt{- НІКОЛИ не замінюй слова синонімами і не змінюй форми слів на альтернативні
  (виказував, кучею, достойною, вивести — залишай як є).}}\\
  \foreignlanguage{ukrainian}{\texttt{- НЕ переформатовуй діалоги: якщо діалог оформлений лапками ("/«»), зберігай
  лапки, НЕ замінюй їх на тире.}}\\
  \foreignlanguage{ukrainian}{\texttt{- НЕ змінюй граматичні форми, які є допустимими варіантами: відмінкові форми
  прикметників (недоступним/недоступний), варіанти дієслів (навчались/навчалися), форми
  займенників (їх/їхній) — якщо форма граматично допустима, залишай її.}}\\
  \foreignlanguage{ukrainian}{\texttt{- Стиль та тон тексту: неформальний текст (чати, смс) залишай як є — не додавай
  крапки в кінці, не прибирай смайлики )).}}\\
  \foreignlanguage{ukrainian}{\texttt{- Порядок слів у реченні.}}\\
  \foreignlanguage{ukrainian}{\texttt{- Великі/малі літери, крім початку речення після крапки.}}\\
  \foreignlanguage{ukrainian}{\texttt{- Лапки: зберігай оригінальний стиль.}}\\
  \foreignlanguage{ukrainian}{\texttt{- Розділові знаки кінця речення: НЕ змінюй . на ? або навпаки.}}\\
  \foreignlanguage{ukrainian}{\texttt{- НЕ додавай тире (—) там, де його не було в оригіналі, окрім прямої мови.}}\\
  \foreignlanguage{ukrainian}{\texttt{- Дефіс у складених словах та повторах (міцно-міцно, дере-дере-дере, все-таки, все
  ж таки — залишай як є).}}\\
  \foreignlanguage{ukrainian}{\texttt{- Якщо сумніваєшся — НЕ змінюй.}}\\

  \foreignlanguage{ukrainian}{\texttt{Приклади:}}\\

  \foreignlanguage{ukrainian}{\texttt{Вхід: Так само потерпає Україна і сьогодні від того що насправді талановитим людям
  заважають працювати усілякі посередності "у руля".}}\\
  \foreignlanguage{ukrainian}{\texttt{Вихід: Так само потерпає Україна і сьогодні від того, що насправді талановитим
  людям заважають працювати усілякі посередності "у руля".}}\\

  \foreignlanguage{ukrainian}{\texttt{Вхід: Це пов'язано з тим, що такі колективні рухи молекул води сильно збільшують
  характерні часи процесів які відбуваються в системі.}}\\
  \foreignlanguage{ukrainian}{\texttt{Вихід: Це пов'язано з тим, що такі колективні рухи молекул води сильно збільшують
  характерні часи процесів, які відбуваються в системі.}}\\

  \foreignlanguage{ukrainian}{\texttt{Вхід: Я ніколи не навчався у медичному коледжі, кажу, - Я лиш підійшов як звичайна
  людина подивитись чи можу бути чимось корисний.}}\\
  \foreignlanguage{ukrainian}{\texttt{Вихід: Я ніколи не навчався у медичному коледжі, кажу, - Я лиш підійшов як
  звичайна людина подивитись, чи можу бути чимось корисний.}}\\

  \foreignlanguage{ukrainian}{\texttt{Вхід: Це у місті швидка приїжджає, забирає хворого і везе у лікарню; якщо ж до
  лікарні кілька годин льоту то хворого можна і не довезти.}}\\
  \foreignlanguage{ukrainian}{\texttt{Вихід: Це у місті швидка приїжджає, забирає хворого і везе у лікарню; якщо ж до
  лікарні кілька годин льоту, то хворого можна і не довезти.}}\\

  \foreignlanguage{ukrainian}{\texttt{Вхід: Найбільше він любив тримати в руках старанно орнаментовані стародавні шолом
  і меч, котрі своїм золотом, здається гріли старого.}}\\
  \foreignlanguage{ukrainian}{\texttt{Вихід: Найбільше він любив тримати в руках старанно орнаментовані стародавні шолом
  і меч, котрі своїм золотом, здається, гріли старого.}}\\

  \foreignlanguage{ukrainian}{\texttt{Вхід: - Я часто казав тобі, що ти дурненька, - сказав він.}}\\
  \foreignlanguage{ukrainian}{\texttt{Вихід: — Я часто казав тобі, що ти дурненька, — сказав він.}}\\

  \foreignlanguage{ukrainian}{\texttt{Вхід: Така традиція також походить з Візантії, прикладом є зображення Андроніка II
  Палеолога.}}\\
  \foreignlanguage{ukrainian}{\texttt{Вихід: Така традиція також походить із Візантії, прикладом є зображення Андроніка
  II Палеолога.}}\\

  \foreignlanguage{ukrainian}{\texttt{Вхід: Як і більшість ділових людей, він не знав напамять жодного вірша і не
  памятав жодної казки, а тому щоразу мусив імпровізувати.}}\\
  \foreignlanguage{ukrainian}{\texttt{Вихід: Як і більшість ділових людей, він не знав напам'ять жодного вірша і не
  памятав жодної казки, а тому щоразу мусив імпровізувати.}}\\

  \foreignlanguage{ukrainian}{\texttt{Вхід: Настя, привіт! я хотіла уточнити про завдання Олі.}}\\
  \foreignlanguage{ukrainian}{\texttt{Вихід: Насте, привіт! Я хотіла уточнити про завдання Олі.}}\\

  \foreignlanguage{ukrainian}{\texttt{Вхід: Я принесу твої улюблені солодощі та обніму тебе міцно-міцно.}}\\
  \foreignlanguage{ukrainian}{\texttt{Вихід: Я принесу твої улюблені солодощі та обніму тебе міцно-міцно.}}\\

  \foreignlanguage{ukrainian}{\texttt{Вхід: Смакота ще та, скажу я вам))}}\\
  \foreignlanguage{ukrainian}{\texttt{Вихід: Смакота ще та, скажу я вам))}}\\

  \foreignlanguage{ukrainian}{\texttt{Вхід: У той час глибокий сенс народної мудрості нам був ще недоступним через брак
  досвіду.}}\\
  \foreignlanguage{ukrainian}{\texttt{Вихід: У той час глибокий сенс народної мудрості нам був ще недоступним через брак
  досвіду.}}\\

  \foreignlanguage{ukrainian}{\texttt{Вхід: Каналізація в будинках еволюціонує повільно, але все ж таки змінюється.}}\\
  \foreignlanguage{ukrainian}{\texttt{Вихід: Каналізація в будинках еволюціонує повільно, але все ж таки змінюється.}}\\

  \foreignlanguage{ukrainian}{\texttt{Вхід: "Досить добре вийшов з Весту, чи не так?" - спитав міліціонер.}}\\
  \foreignlanguage{ukrainian}{\texttt{Вихід: "Досить добре вийшов з Весту, чи не так?" — спитав міліціонер.}}\\

  \foreignlanguage{ukrainian}{\texttt{Вхід: Не знаю як у інших, а у мене в житті траплялось не так багато див.}}\\
  \foreignlanguage{ukrainian}{\texttt{Вихід: Не знаю, як у інших, а у мене в житті траплялось не так багато див.}}\\

  \foreignlanguage{ukrainian}{\texttt{Вхід: Хочеться закінчити у дусі книг з самодопомоги.}}\\
  \foreignlanguage{ukrainian}{\texttt{Вихід: Хочеться закінчити у дусі книг із самодопомоги.}}\\

  \foreignlanguage{ukrainian}{\texttt{Вхід: Наступного дня тим же автобаном повернулися назад, звідти – ще півтори
  години літаком.}}\\
  \foreignlanguage{ukrainian}{\texttt{Вихід: Наступного дня тим же автобаном повернулися назад, звідти — ще півтори
  години літаком.}}\\

  \foreignlanguage{ukrainian}{\texttt{Вхід: От читаєш такі новини і жалкуєш, що населенню України Господь дав все, крім
  совісті і мозгів.}}\\
  \foreignlanguage{ukrainian}{\texttt{Вихід: От читаєш такі новини і жалкуєш, що населенню України Господь дав усе, крім
  совісті і мозгів.}}\\

  \foreignlanguage{ukrainian}{\texttt{Поверни ЛИШЕ виправлений текст.}} \textit{(`Return ONLY the corrected text.')}

\end{quote}

\noindent\textit{English: The instruction structure is the same as optimized-v1 (\ref{prompt:apo A002}); see the
translation there. This version includes additional few-shot examples (lines 6--16 above) and was further refined on
Gemini~3-Flash. Keywords: \foreignlanguage{ukrainian}{\texttt{Виправляй}}~=~`Fix';
\foreignlanguage{ukrainian}{\texttt{НЕ змінюй}}~=~`Do NOT change';
\foreignlanguage{ukrainian}{\texttt{Приклади}}~=~`Examples';
\foreignlanguage{ukrainian}{\texttt{Вхід}}/\foreignlanguage{ukrainian}{\texttt{Вихід}}~=~`Input'/`Output'.}

% ============================================================

\clearpage                                                                                                         
  \section{Inference Pipeline and Structured Output}                                                            
  \label{app:inference-pipeline}
                 
  Each input sentence is processed independently through a single LLM
  call, with no cross-sentence batching. Before any model invocation,                                           
  the pipeline applies a lightweight passthrough rule: lines matching
  the document-marker pattern \verb|# <digits>| (e.g.\ \texttt{\# 0001})                                        
  are emitted verbatim and never sent to the LLM. These markers
  delimit document boundaries in the UA-GEC corpus and carry no                                                 
  correctable content, so routing them around the model both saves
  tokens and prevents spurious edits.                                                                           
                  
  For all remaining sentences, the agent issues one chat completion                                             
  through a LiteLLM router that abstracts over the underlying provider
  (OpenAI, Anthropic, Google, Moonshot). The system message is the                                              
  configured prompt template; the user message is the raw source                                                
  sentence. To eliminate free-form post-processing of model output, we
  constrain the response with a JSON schema derived from a Pydantic                                             
  model and attached to the request as a strict
  \texttt{response\_format}. Field descriptions declared on the                                                 
  Pydantic model propagate into the schema and act as in-band
  instructions to the model.                                                                                    
                  
  \paragraph{Schema definition.}                                                                                
  The response contract is declared once as a Pydantic class:
             
  \begin{lstlisting}[language=Python,basicstyle=\ttfamily\small,frame=single]                                   
  class GECResponse(BaseModel):
      corrected_sentence: str = Field(                                                                          
          ..., description="Corrected version of the input sentence")
  \end{lstlisting}

  \paragraph{Generated JSON schema.}                                                                            
  At call time, the class is converted to JSON Schema, all object nodes
  are closed with \texttt{additionalProperties: false}, and the result                                          
  is wrapped into the provider-agnostic \texttt{response\_format}
  envelope:                                                                                                     
   
  \begin{lstlisting}[basicstyle=\ttfamily\small,frame=single]                                                   
  {               
    "type": "json_schema",
    "json_schema": {                                                                                            
      "name": "GECResponse",
      "strict": true,                                                                                           
      "schema": { 
        "type": "object",
        "additionalProperties": false,
        "required": ["corrected_sentence"],
        "properties": {                                                                                         
          "corrected_sentence": {
            "type": "string",                                                                                   
            "description": "Corrected version of the input sentence"
          }
        }
      }                                                                                                         
    }
  }                                                                                                             
  \end{lstlisting}
 \begin{figure}[h]                                                                                             
  \centering                                                                                                  
  \small                                                                                                        
  \begin{tikzpicture}[                                                                                        
      node distance=5mm,                                                                                        
      every node/.style={font=\footnotesize},                                                                   
      box/.style={rectangle, draw, rounded corners=2pt,                                                         
                  align=center, inner sep=3pt, minimum width=42mm},                                             
      decision/.style={diamond, draw, aspect=2.4, align=center,                                                 
                       inner sep=0pt, minimum width=42mm},                                                      
      side/.style={rectangle, draw, rounded corners=2pt,                                                        
                   align=center, inner sep=3pt, minimum width=28mm,                                             
                   fill=gray!10},                                                                               
      arrow/.style={-{Latex[length=1.6mm]}, semithick},                                                         
  ]    
  \node[box]                       (src)    {source sentence};                                                  
  \node[decision, below=of src]    (mark)   {marker \texttt{\# NNNN}?};
  \node[side, right=8mm of mark]   (pass)   {emit verbatim\\(no LLM call)};                                     
  \node[box, below=of mark]        (call)   {\texttt{router.completion(\dots)}\\                                
                                            structured \texttt{response\_format}};                              
  \node[box, below=of call]        (router) {LiteLLM Router\\                                                   
                                            $\rightarrow$ provider};                                            
  \node[box, below=of router]      (val)    {\texttt{GECResponse.model\_validate}};
  \node[box, below=of val]         (out)    {\texttt{predictions.eval.txt}};               
  \draw[arrow] (src)    -- (mark);                                                                              
  \draw[arrow] (mark)   -- node[above, font=\scriptsize]{yes} (pass);                                           
  \draw[arrow] (mark)   -- node[right, font=\scriptsize]{no}  (call);                                           
  \draw[arrow] (call)   -- (router);                                                                            
  \draw[arrow] (router) -- (val);                                                                               
  \draw[arrow] (val)    -- (out);                                                 
  \end{tikzpicture}
            
  \vspace{2mm}    

  \begin{tabular}{@{}ll@{}}                                                                                     
  \toprule
  \textbf{Parameter} & \textbf{Value (from run YAML)} \\                                                        
  \midrule                                                                                                      
  \texttt{model}             & \texttt{gpt-4.1-mini} \\
  \texttt{temperature}       & \texttt{0.0} \\                                                                  
  \texttt{top\_p}            & \texttt{0.1} \\
  \texttt{reasoning\_effort} & \texttt{None} \\                                                                 
  \texttt{timeout}           & \texttt{90} s \\                                                                 
  \texttt{response\_format}  & \texttt{json\_schema(GECResponse)} \\
  \bottomrule                                                                                                   
  \end{tabular}   
   
  \caption{Per-sentence inference flow. Document markers bypass the LLM;                                        
  all other sentences go through one structured-output call. Decoding
  parameters are read verbatim from the run YAML, ensuring                                                      
  deterministic replay.}
  \label{fig:inference-flow}                                                                                    
  \end{figure}          
  The decoding parameters in Figure~\ref{fig:inference-flow} are read                                           
  from the run YAML and the exact configuration file is copied into
  the output directory alongside \texttt{results.json}, so a run can be                                         
  replayed bit-for-bit given the same provider model snapshot.                                                  
    The returned payload is validated with                                             
  \texttt{model\_validate}, so any schema violation is caught                                                   
  deterministically rather than being masked by string heuristics. If a                                         
  provider rejects \texttt{json\_schema}, the client transparently
  retries with \texttt{response\_format = \{"type":"json\_object"\}};                                           
  for the single-field case, a final recovery path extracts the                                                 
  corrected sentence from malformed JSON to keep evaluation aligned.
  This design ensures that every non-marker sentence yields exactly one                                         
  validated correction, making the sentence-to-prediction mapping
  bijective and the run reproducible given a fixed configuration.                                    
\end{document}